\documentclass[11pt]{article}

\usepackage[final]{acl}

\PassOptionsToPackage{table}{xcolor}
\usepackage[table]{xcolor}
\usepackage{colortbl}
\definecolor{myblue}{rgb}{0.88, 0.94, 0.95}

\usepackage{float}
\usepackage{makecell}
\usepackage{booktabs}
\usepackage{graphicx}
\usepackage{multirow}
\usepackage{multicol}
\usepackage{amsmath}
\usepackage{amssymb}
\usepackage{arydshln}
\usepackage{pifont}
\usepackage{array}
\usepackage{amsmath}
\usepackage{amssymb}
\usepackage{algpseudocode}
\usepackage{listings}
\usepackage{algorithm}
\usepackage{alltt}
\usepackage{inconsolata}
\usepackage{subcaption} \usepackage{placeins}
\usepackage{times}
\usepackage{latexsym}
\usepackage{xspace}
\usepackage{url, hyperref}
\usepackage{xcolor}
\usepackage{soul}
\sethlcolor{cyan!30}
\newcommand{\hlcyan}[1]{\colorbox{cyan!30}{#1}}
\newcommand{\hlpink}[1]{\colorbox{magenta!40}{#1}}
\usepackage[most]{tcolorbox}
\newcommand{\ie}{i.e,\xspace}

\definecolor{forestgreen}{rgb}{0.13, 0.55, 0.13}
\definecolor{fireenginered}{rgb}{0.81, 0.09, 0.13}
\definecolor{darkred}{rgb}{0.6, 0.1, 0.1}
\newcommand{\cmark}{{\textcolor{forestgreen}{\ding{51}}}}\newcommand{\xmark}{{\textcolor{darkred}{\ding{55}}}}\definecolor{veloblue}{RGB}{83, 149, 218}
\definecolor{veloorange}{RGB}{236, 133, 45}
\definecolor{lightgreen}{HTML}{E6F4EA}
\definecolor{green}{HTML}{C6EFCE}
\definecolor{LightGray}{gray}{0.95}
\definecolor{lightyellow}{HTML}{FFEBED}
\usepackage[percent]{overpic}

\newcommand{\visual}{visual cue window\xspace}
\newcommand{\spoken}{spoken narration window\xspace}
\newcommand{\paired}{spoken-visual window\xspace}
\newcommand{\cmdad}{CMD-AD\xspace}
\newcommand{\madeval}{MAD-Eval\xspace}
\newcommand{\longlsmdc}{LongLSMDC\xspace}
\newcommand{\ours}{Cue2Narrate\xspace}
\newcommand{\stageone}{\underline{Cue}2Narrate\xspace}
\newcommand{\stagetwo}{Cue2\underline{Narrate}\xspace}

\usepackage[table]{xcolor}
\definecolor{ourrow}{RGB}{220, 235, 250}  \usepackage[most]{tcolorbox}

\newcommand{\promptbox}[2]{  \begin{center}
    \begin{tcolorbox}[
                float, floatplacement=tp,              breakable,
        enhanced jigsaw,
        width=0.95\linewidth,         top=0.3em,
        bottom=0.3em,
        left=0.5em,
        right=0.5em,
        toptitle=0.3em,
        bottomtitle=0.2em,
        boxsep=0pt,
                        colback=gray!4!white,
        colframe=black!55!white,
        colbacktitle=black!65!white,
        coltitle=white,
        fonttitle=\bfseries\small,
        arc=2pt,
        shadow={1.5pt}{-1.5pt}{0pt}{black!15},
        boxrule=0.5pt,
        title={\footnotesize\textbf{#1}}
    ]
      \footnotesize
      #2
    \end{tcolorbox}
  \end{center}
}
\newcommand{\userpromptbox}[2]{  \noindent
  \begin{tcolorbox}[
      breakable,
      enhanced jigsaw,
      width=\columnwidth,
      top=0.3em,
      bottom=0.3em,
      left=0.5em,
      right=0.5em,
      toptitle=0.3em,
      bottomtitle=0.2em,
      boxsep=0pt,
      colframe=blue!70!black,
      colback=blue!5,
      boxrule=0.5pt,
      title={\footnotesize\textbf{#1}}
  ]
    \footnotesize
    #2
  \end{tcolorbox}}
\usepackage{tikz}
\usetikzlibrary{calc}

\usepackage[T1]{fontenc}

\usepackage[utf8]{inputenc}

\usepackage{enumitem}

\title{From Visual Cues to Spoken Narration: Rethinking Audio Description}

\author{
\textbf{Akshita Gupta}$^{1}$ \quad
\textbf{Aditya Arora}$^{1}$ \quad
\textbf{Federico Tombari}$^{2,3}$ \\
\textbf{Marcus Rohrbach}$^{1}$ \quad
\textbf{Anna Rohrbach}$^{1}$ \\
\\
$^{1}$TU Darmstadt \& hessian.AI, Darmstadt, Germany \\
$^{2}$Google Research \quad
$^{3}$TU Munich \\
\texttt{akshita.gupta@tu-darmstadt.de}
}

\begin{document}

\maketitle

\begin{abstract}
Audio Description (AD) provides spoken narration of visual events during dialogue gaps, making movies accessible 
to visually impaired audiences.
The problem requires determining both \emph{what} (which visual event) and \emph{when} (position for inserting the AD) to narrate, to achieve the best user experience. Prior work has largely reduced the problem to video captioning of pre-segmented video clips, i.e., \emph{what} is largely predefined and \emph{when} is ignored entirely.
We propose \ours, a two-stage pipeline that jointly predicts \emph{what} and \emph{when} to narrate in longer untrimmed movie clips. A dual-head audio-visual localizer predicts two temporally distinct windows per AD utterance: a \visual and a \spoken.
A LoRA-adapted VLM then generates concise ADs from the predicted visual 
evidence, trained with a Description Ranking Loss that ranks captions (negative samples) of the same frames lower than the GT AD. To benchmark this new problem statement, we introduce the \longlsmdc benchmark with up to 8-min movie clips ($\sim$6.5\,min on average). On \longlsmdc, \ours outperforms video-only and audio-only localization baselines by 5--12 points in avg. mAP. Under both predicted- and GT-window evaluation, \ours improves AD generation over the corresponding fine-tuned base VLM. These results establish the first benchmark for multi-segment AD generation on long-form clips. Data \& Code: \href{https://github.com/multimodal-ai-lab/Cue2Narrate}{multimodal-ai-lab/Cue2Narrate}
\end{abstract}    

\begin{figure*}[!t]
\centering
\includegraphics[width=\textwidth]{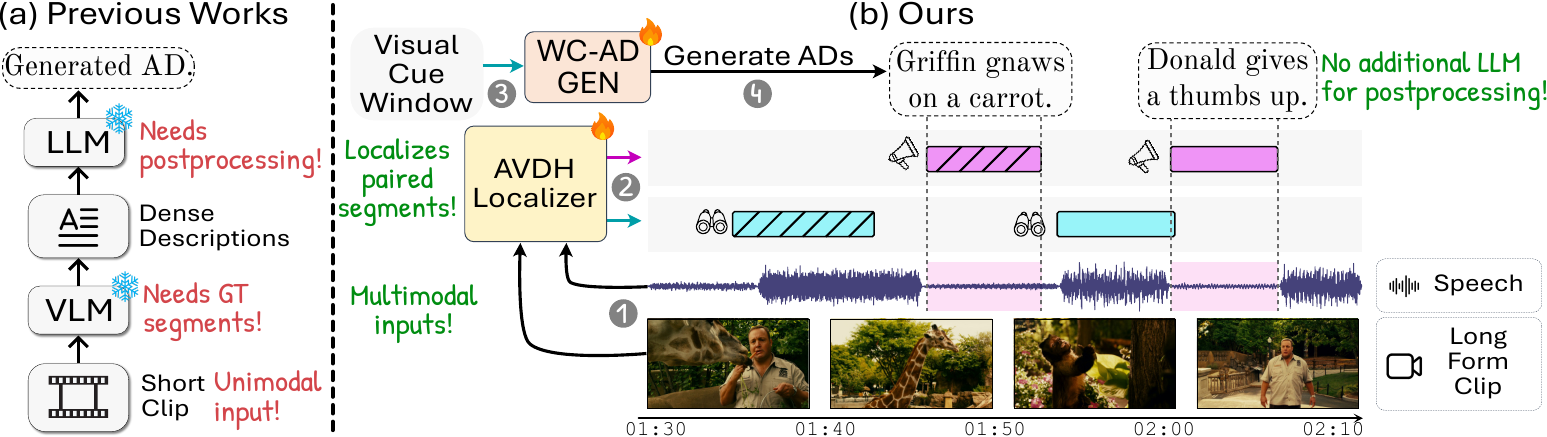}
\caption{
\textbf{\ours overview.}
\textbf{(a)} Prior methods caption GT clips with a VLM and then refine the dense descriptions with an LLM into AD-style text.
\textbf{(b)} Given a long-form clip and the corresponding audio, our AVDH Localizer (\stageone) predicts paired temporal windows per AD: a \hlcyan{\visual} marking when the event is on screen, and a \hlpink{\spoken} marking the dialogue gap in which the AD is inserted. 
Our WC-AD Gen (\stagetwo) generates a description from the localized visual window, to be inserted into the predicted spoken window.
The figure shows an actual example from \longlsmdc, where these windows are routinely offset.}
\label{fig:teaser}
\end{figure*}

\section{Introduction}
\label{sec:intro}

Audio Description (AD) is spoken narration that makes movies accessible to blind and low-vision audiences by describing on-screen content not conveyed through dialogue~\cite{benecke2007audio}. A trained human describer watches the film, identifies the visual events that matter to the story (actions, expressions, scene changes, character interactions), and writes short descriptions to be voiced over the soundtrack~\cite{packer2015overview, dcmp2020descriptionkey, pavel2020rescribe}. Since AD is mixed into the existing audio, each description must ideally be inserted into a dialogue ``gap'', avoiding overlap with speech~\cite{pavel2020rescribe}. The describer therefore makes two distinct decisions each time: \emph{what} visual event to describe, and \emph{when} the description should be voiced. The latter is typically done in the
gap that occurs during or slightly before/after the event itself~\cite{pavel2020rescribe, branje2012livedescribe, fryer2016introduction}. However, professional AD production is slow and expensive, costing
roughly \$15--\$75 per finished minute, limiting 
accessibility to a small fraction of released films~\cite{Whitney2018ADCost,3playmedia_ad_vendor}; 
automating this process is essential to scale AD coverage.
Most prior works on automatic AD generation significantly reduce the problem. First, they often evaluate in an oracle-window setting: the model is given the temporal interval for which a description should be produced~\cite{Rohrbach2017LSMDC,Han2023AutoAD,han2024autoad,Fang2025DistinctAD,Xie2025ShotByShot}. Namely, AD boils down to video captioning over pre-defined short clips. Besides, prior works ignore the distinction between the \emph{what} and \emph{when} intervals. Existing AD datasets provide one type of temporal windows: CMD-AD and TV-AD~\cite{han2024autoad,Xie2024AutoADZero} align AD text to AudioVault\footnote{AudioVault is a community-maintained non-profit repository of movie ADs. See \url{https://audiovault.net}.}
\emph{spoken} slots, while LSMDC and MAD-Eval~\cite{Rohrbach2017LSMDC,Han2023AutoAD} provide \emph{visually}-aligned short clips with text. Prior works do not highlight this distinction. To sum up, most evaluate \emph{what to say} given an oracle window without the challenge of long untrimmed video, and without predicting \emph{when to actually speak}. The few works that touch on description timing inherit these limitations: AutoAD-II~\cite{Han2023AutoAD2} binary-classifies pre-supplied speech gaps, and CA3D~\cite{Lee2025CA3D} detects a single event per short clip. Thus, the full AD problem, with visual/spoken cue windows in untrimmed video with multi-segment prediction, has not been tackled so far.

We address these limitations twofold. First, we introduce \textbf{\longlsmdc}, a long-form AD benchmark with up to 8-min movie clips ($\sim$6.5\,min on average)
with dense multi-segment annotations, built on top of LSMDC~\cite{Rohrbach2017LSMDC,lsmdcv2}. 
Unlike prior benchmarks, \longlsmdc provides both types of windows per AD utterance, matching how human describers produce AD: the \visual is paired with the \spoken in which it is voiced. 
Second, we propose \textbf{\ours} (Fig.~\ref{fig:teaser}), a paired-window AD pipeline for untrimmed video: a dual-head audio-visual localizer (AVDH Localizer) predicts both windows (visual and spoken) per AD utterance, and WC-AD generator produces the description from the predicted localized \visual, to be inserted into the \spoken. 
On \longlsmdc, our localizer outperforms a video-only ActionFormer~\cite{Zhang2022ActionFormer} baseline by 11.8 points in avg. mAP, an audio-only ActionFormer variant by 5.0 points, and a dual-head ActionFormer baseline 
by 3.6 points. We also show transfer to \cmdad and \madeval, where we evaluate on the respective temporal windows.
In terms of AD quality, \ours improves over its base VLM on both \longlsmdc and \cmdad under both predicted- and GT-window evaluation.
Generated ADs are matched to GT via tIoU on the predicted spoken-narration windows, and to our knowledge ours is the first protocol
to jointly evaluate \emph{what} and \emph{when} in AD generation.

\begin{table*}[t]
\centering
\small
\setlength{\tabcolsep}{4pt}
\caption{\textbf{Audio description datasets} compared across the dimensions needed to predict both AD windows on long-form clips. Prior datasets are either short-clip (LSMDC, MAD-Eval) or contain spoken-only segments on longer clips (CMD-AD, TV-AD; frames available by source). \longlsmdc is the first to support multi-segment prediction of the visual and spoken windows on long-form clips.}
\begin{tabular}{@{}lllrcccc@{}}
\toprule
Dataset & Source & Domain & \#Annotations & Clip length & ADs/clip & Visual & Spoken \\
\midrule
LSMDC~\cite{Rohrbach2017LSMDC}     & LSMDC                 & Movies & 108.5K & $\sim$5\,sec & 1         & \cmark & \xmark \\
MAD-Eval~\cite{Han2023AutoAD}      & MAD (LSMDC subset)    & Movies & 7K     & $\sim$5\,sec & 1         & \cmark & \xmark \\
CMD-AD~\cite{han2024autoad}        & CMD + AudioVault      & Movies & 101K   & $\sim$2\,min  & $\sim$10  & \xmark & \cmark \\

TV-AD~\cite{Xie2024AutoADZero}\textsuperscript{\textdagger}     & TV + AudioVault       & TV     & 34K    & unknown       & --        & \xmark & \cmark \\
\midrule

\rowcolor{lightgreen}
\longlsmdc (ours)                  & LSMDC full movies   & Movies & 83.8K  & $\sim$6.5\,min  & $\sim$40  & \cmark & \cmark \\
\bottomrule
\end{tabular}
\\[0.5em]
{\footnotesize \textsuperscript{\textdagger} TV-AD releases only pre-extracted frames, so clip-length statistics cannot be computed.}

\label{tab:datasets}
\end{table*}

\section{Related Work}
\label{sec:related_work}

\textbf{Audio Description Generation.}

Automated AD generation has been studied almost entirely as a text-generation problem given a temporal window. AutoAD~\cite{Han2023AutoAD} introduces VLMs for AD generation from a known interval, leveraging text-only AD corpora and visual captioning datasets for pretraining; AutoAD-II~\cite{Han2023AutoAD2} extends this with character recognition and binary classification of ASR-extracted speech gaps to decide where AD should be inserted, pretraining on AudioVault-AD text and WebVid video-text~\cite{bain2021frozen}; AutoAD-III~\cite{han2024autoad} releases CMD-AD with AudioVault-aligned spoken slots and introduces HowToAD-style long-video pretraining. Subsequent work focuses on richer generation conditioned on the assumed window, often with external character banks or movie scripts: MM-Narrator~\cite{MMNarrator2024}, NarrAD~\cite{Park2025NarrAD}, UniAD~\cite{Wang2025UniAD}, AutoAD-Zero~\cite{Xie2024AutoADZero}, Shot-by-Shot~\cite{Xie2025ShotByShot}, and DistinctAD~\cite{Fang2025DistinctAD}. The corresponding benchmarks preserve only one part of the ``paired'' AD annotation: CMD-AD and TV-AD~\cite{han2024autoad,Xie2024AutoADZero} contain \emph{spoken} slots without a \visual, while LSMDC and MAD-Eval~\cite{Rohrbach2017LSMDC,Han2023AutoAD} contain \emph{visual} slots without \spoken. Parallel work~\cite{kala2025you} also operate on
few-minute video segments rather than trimmed clips, but focus on QA-based evaluation of \emph{generated} ADs rather than predicting AD timing. None support joint prediction 
, and the closest timing-related work, CA3D~\cite{Lee2025CA3D}, predicts a single window per clip rather than a paired structure. We instead predict the pairs in our Stage 1 on long-form clips by using video and audio modalities, supervised by \longlsmdc annotations. Where prior methods generate from an assumed or oversampled visual context, we learn to predict the \visual and pair the generated description with the predicted \spoken, mirroring how human describers produce AD.

\noindent\textbf{Temporal Localization and Dense Captioning.}
Temporal action localization (TAL) identifies multi-segment temporal boundaries in untrimmed video with transformer-based architectures~\cite{Zhang2022ActionFormer,Shi2023TriDet} and cross-modal cues~\cite{tian2018audio,senocak2018learning,senocak2023event}. Dense video captioning extends this to producing descriptions per segment, either as separate~\cite{krishna2017dense} or end-to-end~\cite{iashin2020multi,Li_2018_CVPR,yang2023vid2seq,Islam2022ViS4mer} pipelines. Both lines predict visual-only segments anchored to scene changes or action boundaries, with no requirement on the audio track. AD differs structurally: the output is a paired (visual, spoken) annotation, where the spoken window is a dialogue gap that may sit before, during, or after the event being described~\cite{pavel2020rescribe}. The visual frame at an AD's spoken window therefore need not match the description's text, placing AD outside the scope of visual-only localization and captioning. Our audio-visual dual-head localizer predicts both windows of the pair, with each head supervised against its own ground-truth window.
\section{LongLSMDC Dataset}
\label{sec:dataset}

Our \longlsmdc benchmark is constructed from the full-length movies in LSMDC/LSMDC v2~\cite{Rohrbach2017LSMDC,lsmdcv2}. The commonly used annotations only provide timestamps for \visual{s}. Importantly, we were able to get access to the corresponding \spoken{s}.\footnote{Obtained from the original authors.} We segment each movie into non-overlapping clips of up to 8 min ($\sim$6.5\,min), assign each AD annotation to the clip containing its temporal midpoint, and convert movie-level timestamps to clip-relative coordinates. The resulting dataset contains $1{,}995$ training clips and $170$ validation clips, with an average of ${\sim}40$ AD annotations per clip 
(Table~\ref{tab:datasets}). Training annotations span three supervision categories: $37{,}302$ \paired ($y^{vn}$ and $y^{sp}$), $21{,}295$ visual-only ($y^{vn}$), and $20{,}197$ spoken-only ($y^{sp}$) ($78.8$k training annotations; $83.8$k including validation, Table~\ref{tab:datasets}).\footnote{This is due to various post-processing steps that occured between extraction of the spoken windows and manual alignment to the visual windows carried out in LSMDC.}
Within the paired set, $32{,}715$ (${\sim}88\%$) exhibit a temporal offset between the two windows, confirming that the visual and spoken windows are routinely distinct. The validation set contains $5{,}006$ paired annotations, supporting evaluation against both windows.
We additionally create a per-clip character bank, and action-event annotations from each AD sentence via a closed-source model~\cite{comanici2025gemini} to support the Action Score metric~\cite{Xie2025ShotByShot}. More details are provided in Appendix~\ref{sec:dataset_prep}.

\begin{figure*}[t]
    \centering
    \includegraphics[width=\textwidth]{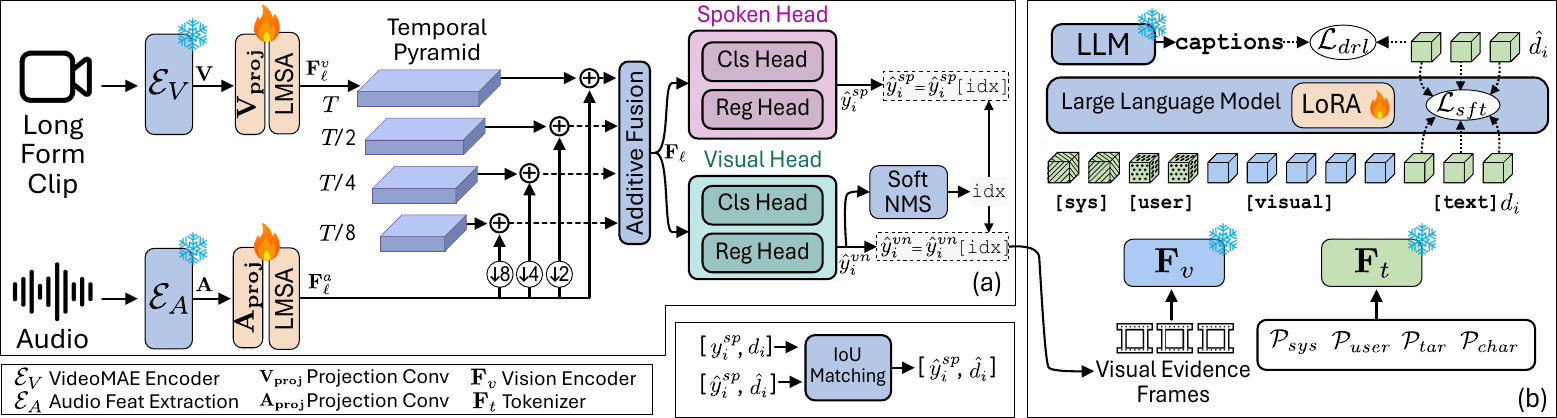}
    \caption{\textbf{\ours architecture.}
    \textbf{(a) Localization (Stage 1).} Frozen video ($\mathcal{E}_V$) and audio
    ($\mathcal{E}_A$) encoders extract features that are projected, refined with
    self-attention (LMSA), and encoded into a 4-level temporal pyramid ($T,
    T/2, T/4, T/8$). The streams are combined by additive fusion. Two
    independent heads the Spoken Head and Visual Head, each with classification and regression branches predict
    the \spoken and \visual, respectively. Soft-NMS yields final segments;
    predicted spoken windows ($\hat{y}_i^{sp}[\texttt{idx}]$) drive evaluation matching,
    while predicted visual windows ($\hat{y}_i^{vn}[\texttt{idx}]$) condition Stage~2.
    \textbf{(b) Generation (Stage 2).} For each predicted \visual, visual
    evidence frames are encoded into $\mathbf{F}_v$. A structured prompt
    $\mathcal{P} = [\mathcal{P}_{sys}, \mathcal{P}_{user}, \mathcal{P}_{target},
    \mathcal{P}_{char}]$ fed to tokenizer $\mathbf{F}_t$. A frozen
    LLM with LoRA-adapted projection is trained with $\mathcal{L}_{sft}$ on
    ground-truth ADs and $\mathcal{L}_{drl}$ against a caption sampled
    from a frozen LLM on the same frames, generating the AD sentence
    $\hat{d}_i$. \textbf{(c) Evaluation.} Under predicted windows, generated ADs must be matched to ground-truth annotations before scoring; we match on \spoken tIoU because that is the side AD references are aligned to in human annotation practice. }
        \label{fig:architecture}
            \end{figure*}

\section{\ours}
\label{sec:method}

\subsection{Problem Formulation}
\label{sec:problem}
 
Let $X=(\mathcal{V},\mathcal{A})$ denote a long-form clip with video $\mathcal{V}$ and audio $\mathcal{A}$.
The target output is a set of AD annotations
\begin{equation}
\mathcal{Y}=\{(d_i,\, y_i^{vn},\, y_i^{sp})\}_{i=1}^{N},
\end{equation}
where $d_i$ is the AD sentence,
$y_i^{vn}=[vn_s^i,\,vn_e^i]$ is the \visual interval in which the described event is visible on screen,
and $y_i^{sp}=[sp_s^i,\,sp_e^i]$ is the \spoken interval in which $d_i$ is delivered during a dialogue gap.
At test time, both the temporal windows and the sentence content are unknown, as well as the number of AD annotations $N$.
Our method factorizes this problem into two stages.
Stage~1 predicts paired temporal windows $(\hat{y}_i^{vn},\hat{y}_i^{sp})$ from the raw clip $X$.
Stage~2 generates a sentence $\hat{d}_i$ from the visual evidence inside the predicted \visual, $\hat{y}_i^{vn}$.
The final delivered AD is $(\hat{d}_i,\hat{y}_i^{sp})$:
the \visual determines \emph{what} is described, and the \spoken determines \emph{when} that description is inserted.
 
\subsection{Stage~1: Audio-Visual Dual-Head (AVDH) Localizer}
\label{sec:stage1}

Stage~1 localizes AD slots in a long-form clip by predicting, for each instance, a \visual $y_i^{vn}$ and a \spoken $y_i^{sp}$. In \longlsmdc, 88\% of paired annotations exhibit a temporal offset between these two windows; the visible event and the spoken narration differ in onset, duration, or both.
We predict them with two independent heads over a shared audio-visual representation, each supervised against its own GT windows. 
We extract visual features $\mathbf{V}\in\mathbb{R}^{T\times 768}$ with a VideoMAE encoder~\cite{tong2022videomae} and audio features $\mathbf{A}\in\mathbb{R}^{T\times 130}$, a 130-dimensional log-mel spectrogram,
resampled to the visual timeline. Fig.~\ref{fig:architecture}~(a)
The audio stream directly exposes speech activity and silence patterns, critical cues for predicting when a narration slot is available. 
The visual features are projected and refined with temporal self-attention, then encoded into a multi-scale temporal pyramid $\{\mathbf{F}^{v}_{\ell}\}_{\ell=1}^{L}$ via a ConvTransformer backbone, covering AD events of different durations.
The audio features are projected to $d_h{=}256$,
refined with lightweight temporal self-attention, and average-pooled to each pyramid level, yielding $\{\mathbf{F}^{a}_{\ell}\}_{\ell=1}^{L}$.
We fuse the two modalities by pyramid-level addition:
\begin{equation}
\mathbf{F}_{\ell}=\mathbf{F}^{v}_{\ell}+\mathbf{F}^{a}_{\ell}.
\label{eq:fusion}
\end{equation}

This design is motivated by the complementary strengths of the two modalities (Figs.~\ref{fig:stage1_analysis},~\ref{fig:duration_qual}): audio dominates \spoken localization where silence patterns are the primary cue, while video dominates \visual localization where scene changes and character actions matter.
Both heads operate on the fused pyramid $\{\mathbf{F}_\ell\}_{\ell=1}^{L}$ but predict independently.
Each head contains a focal classification branch~\cite{lin2017focal} for foreground/background scoring and a regression branch that predicts non-negative left/right offsets $(\delta_l, \delta_r)$ from each temporal anchor~$t$.
For pyramid level~$\ell$ with stride~$s_\ell$, the predicted interval is
\begin{equation}
[t - \delta_l \cdot s_\ell,\; t + \delta_r \cdot s_\ell].
\label{eq:decode}
\end{equation}
The \visual head predicts $y_i^{vn}$; the \spoken head predicts $y_i^{sp}$.
This allows the model to represent windows with different durations and arbitrary temporal offsets. 
For spoken classification, we add a learned temporal-position prior over normalized clip time, giving the head a coarse prior over where narration slots tend to occur in a clip.
Further, Soft-NMS~\cite{bodla2017soft} suppresses duplicate predictions by down-weighting overlapping candidates by their tIoU rather than removing them.
 
\noindent\textbf{Training objective.} Both heads are trained jointly:
\begin{equation}
\mathcal{L}_{\text{stage1}}
= \mathcal{L}^{vn}_{\text{cls}} + \mathcal{L}^{vn}_{\text{reg}}
+ \lambda_{sp}\!\left(\mathcal{L}^{sp}_{\text{cls}} + \mathcal{L}^{sp}_{\text{reg}}\right),
\label{eq:stage1_loss}
\end{equation}
with $\lambda_{sp}{=}1$.
Both classification terms use sigmoid focal loss~\cite{lin2017focal}; both regression terms use a 1D center-based DIoU loss.
\longlsmdc annotations supervise the heads asymmetrically:
\visual-only annotations train only the visual head,
\spoken-only annotations train only the spoken head,
and \paired annotations train both.
The paired subset is the key supervision that prevents the two heads from collapsing to a single shared window.
 
\subsection{Stage~2: Window-Conditioned AD Generator (WC-AD GEN)}
\label{sec:stage2}
Stage~2 maps a localized \visual to a single-sentence audio description, Fig.~\ref{fig:architecture}~(b).
Beyond describing the correct event, the generator must produce narration in the concise AD style rather than captioning the whole clip in great detail.
Standard cross-entropy fine-tuning teaches the model \emph{what} to say but not \emph{how}: AD training data is too sparse to overwrite the pretraining prior towards long, descriptive captions. 
We address this with a Description Ranking Loss (DRL) that suppresses the caption prior during training.
We build Stage~2 on Qwen models~\cite{Qwen25VL} as the backbone, adapting it with LoRA ($r{=}16$, $\alpha{=}32$, dropout~$0.05$) on the language model and multimodal projection while keeping the vision encoder frozen. Frames are sampled uniformly from the \visual ($F{=}32$ at training, $F{=}64$ at inference) and encoded by the frozen vision encoder, then aligned via a trainable projection layer.
For each annotation~$i$, the model receives a prompt $\mathcal{P}_i$ containing:
(i)~a task instruction for one-sentence AD generation,
(ii)~visual tokens from $y_i^{vn}$,
(iii)~optional character-bank identities, and
(iv)~target-window context.
The \spoken is not used as visual evidence; it determines position of the generated sentence.

\noindent\textbf{Loss functions.}
Let $y_i^{+}{=}d_i$ denote the ground-truth AD and $\mathbf{V}_i$ the sampled frames from $y_i^{vn}$.
The supervised fine-tuning loss is
\begin{equation}
\mathcal{L}_{\text{sft}} = -\sum_t \log P(y^{+}_{i,t} \mid y^{+}_{i,<t},\, \mathbf{V}_i,\, \mathcal{P}_i).
\label{eq:sft}
\end{equation}
We additionally compute a ranking loss to suppress the caption prior.
For each training sample, a negative $y_i^{-}$ is precomputed by prompting a frozen LLM on $\mathbf{V}_i$ with a simple prompt (no AD framing) and cached offline.
We compute mean per-token log-likelihoods for the ground-truth and the negative under the same model and prompt:
\begin{equation}
\begin{aligned}
s_i^{+} &= \tfrac{1}{|y_i^{+}|}\sum_t \log P(y^{+}_{i,t} \mid y^{+}_{i,<t},\, \mathbf{V}_i,\, \mathcal{P}_i), \\
s_i^{-} &= \tfrac{1}{|y_i^{-}|}\sum_t \log P(y^{-}_{i,t} \mid y^{-}_{i,<t},\, \mathbf{V}_i,\, \mathcal{P}_i),
\end{aligned}
\label{eq:scores}
\end{equation}
where prompt tokens and padding positions are excluded by the label mask.
DRL encourages higher normalized likelihood for the concise AD than for the ``vanilla'' caption:
\begin{equation}
\mathcal{L}_{\text{drl}} = -\log\sigma\!\left(\tfrac{s_i^{+} - s_i^{-}}{\tau}\right).
\label{eq:drl}
\end{equation}
Since the negative is generated by the frozen VLM, DRL directly targets the residual pretraining prior that cross-entropy alone cannot suppress.
Thus, the full Stage~2 objective is:
\begin{equation}
\mathcal{L}_{\text{stage2}} = \mathcal{L}_{\text{sft}} + \lambda\,\mathcal{L}_{\text{drl}}, \qquad \lambda{=}0.2,\;\tau{=}0.5.
\label{eq:stage2}
\end{equation}
During training, Stage~2 uses ground-truth \visual s; at inference, predictions from Stage~1 are used instead.
 
\subsection{Joint Inference}
\label{sec:inference}
  
The two stages couple naturally at test time: Stage~1 predicts \emph{where} to look and narrate, Stage~2 generates exactly \emph{what} to say, the same decisions a human describer makes.
At test time, Stage~1 predicts paired windows $\{(\hat{y}_i^{vn}, \hat{y}_i^{sp})\}_{i=1}^{\hat{N}}$ from the raw clip.
For each pair, Stage~2 generates $\hat{d}_i$ from $\hat{y}_i^{vn}$ alone. No previously generated AD is supplied as context ($k=0$); the phrase refers only to standard autoregressive decoding within a single sentence.
The final output is:
\begin{equation}
\{(\hat{d}_i,\, \hat{y}_i^{sp})\}_{i=1}^{\hat{N}},
\end{equation}
with \emph{no ground-truth boundaries used at any point}. The total cost is one Stage~1 forward pass and one VLM decoding per AD slot. 
\section{Experiments}
\label{sec:experiments}

\noindent\textbf{Datasets.}
For Stage~1, we train on \longlsmdc and evaluate on \longlsmdc, CMD-AD~\cite{han2024autoad}, and MAD-Eval~\cite{Han2023AutoAD}. For Stage~2, we only use the datasets with \spoken's annotations. More dataset and metric details in Appendix~\ref{sec:dataset_prep}.

\begin{table}[t]
\begingroup
\centering
\caption{\textbf{Stage~1 localization and transfer.} Average mAP over tIoU $0.2$--$0.6$.
All methods are trained on \longlsmdc. \textbf{Visual} and \textbf{Spoken} report each
head's mAP against its own GT window; \textbf{Avg.} is the mean.
ActionFormer rows are single-head configurations; \emph{Baseline}
is a two-head ActionFormer receiving the same audio-visual input with naive modality
combination. Transfer to \cmdad reports spoken narration window mAP and to \madeval
visual cue window mAP. ZS = zero-shot; FT = fine-tuned on the target dataset.
``--'' indicates not evaluated or not predicted.}
\label{tab:stage1_mAP_by_test}
\scriptsize
\setlength{\tabcolsep}{2.4pt}
\renewcommand{\arraystretch}{1.2}
\resizebox{\columnwidth}{!}{\begin{tabular}{lcccccccc}
\toprule
\multirow{2}{*}{Method} & \multicolumn{2}{c}{Modality} & \multicolumn{3}{c}{\longlsmdc} & \multicolumn{2}{c}{\cmdad} & \madeval\\
\cmidrule(lr){2-3}\cmidrule(lr){4-6}\cmidrule(lr){7-8}\cmidrule(lr){9-9}
& Video & Audio & Visual & Spoken & Avg. & FT & ZS & ZS\\
\midrule
Spectrogram & \xmark & \cmark & 1.4 & 1.3 & 1.4 & 1.7 & 1.7  & 23.1\\
Gemini~2.5~Pro & \cmark & \cmark & 7.1 & 4.4 & 5.8 & -- & -- & --\\
\midrule
ActionFormer & \cmark & \xmark & 32.0 & 28.5 & 30.3 & -- & -- & 63.7\\
Audio-ActionFormer & \xmark & \cmark & 35.2 & 39.1 & 37.1 & 39.2 & 25.0 & --\\
Baseline & \cmark & \cmark & 39.0 & 38.1 & 38.5 & -- & -- & --\\
\midrule
\textbf{\stageone} & \cmark & \cmark & \textbf{42.2} & \textbf{42.0} & \textbf{42.1} & \textbf{41.0} & \textbf{26.5} & \textbf{70.0}\\
\bottomrule
\end{tabular}}
\endgroup
\end{table}

\noindent\textbf{Joint evaluation.} 
\label{sec:joint_eval}
When the AD Generator is conditioned on Stage~1 predictions, Stage~2 generates AD text from predicted \visual , while temporal matching for metrics is performed
on the paired predicted \spoken. Concretely, the predicted \visual
drives AD generation only, while the predicted \spoken drives temporal
matching to GT spoken annotations. A prediction is matchable only against GT
spoken segments from the same video with tIoU $> 0.3$; among multiple matches
to the same GT segment, greedy one-to-one matching retains the highest tIoU. Text metrics (CIDEr~\cite{vedantam2015cider}, Action Score~\cite{Xie2025ShotByShot}, Recall~\cite{Han2023AutoAD},
LLM-Eval~\cite{han2024autoad,MMNarrator2024})
are computed on matched pairs only, reducing false positives from
unmatched localization errors and keeping evaluation aligned with the
spoken-side annotation protocol.

\begin{figure}[t]
  \centering
  \begin{subfigure}[b]{0.49\columnwidth}
    \centering
    \includegraphics[width=\linewidth]{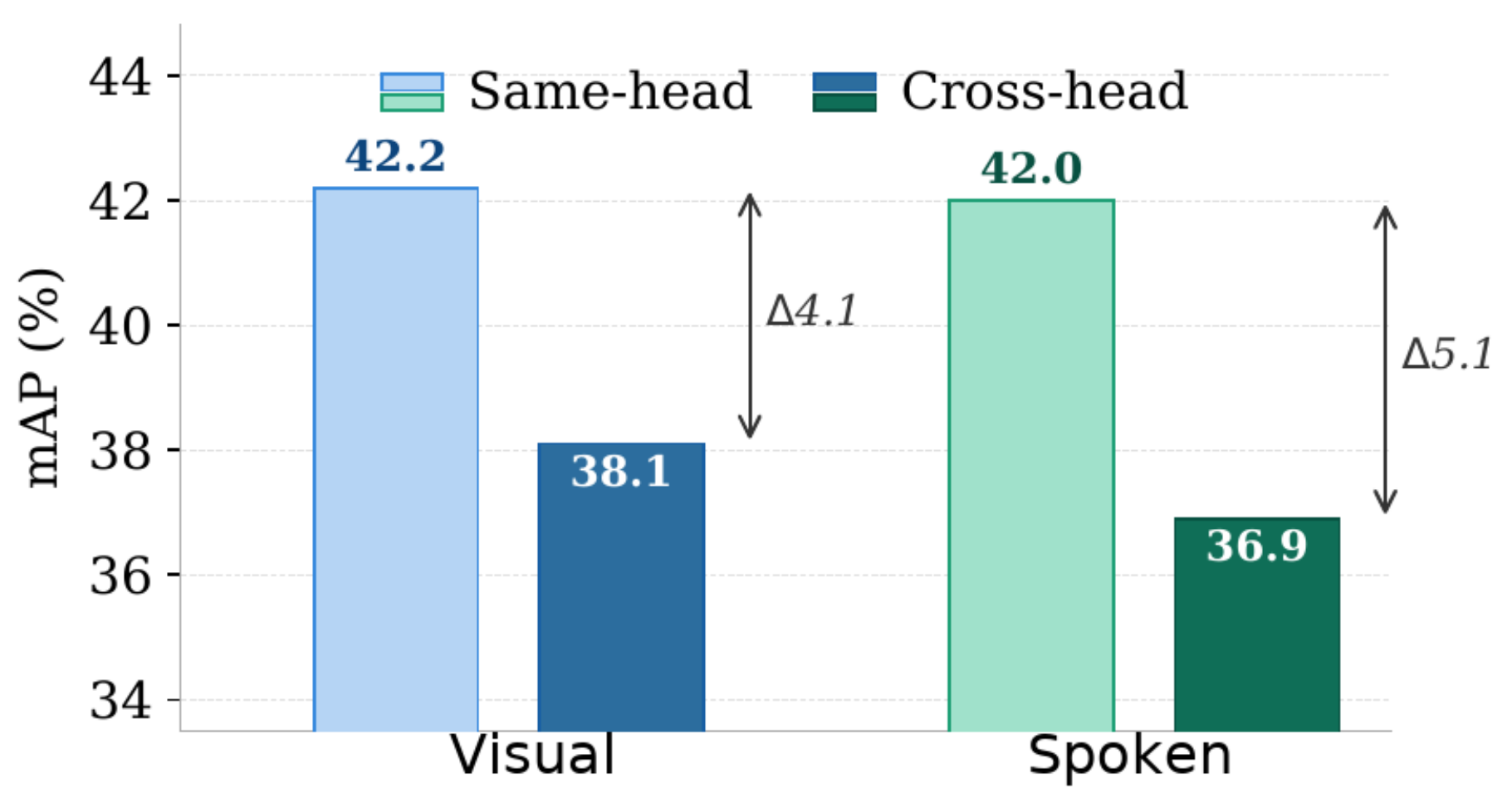}
    \caption{}
    \label{fig:cross_head}
  \end{subfigure}
  \hfill
  \begin{subfigure}[b]{0.49\columnwidth}
    \centering
    \includegraphics[width=\linewidth]{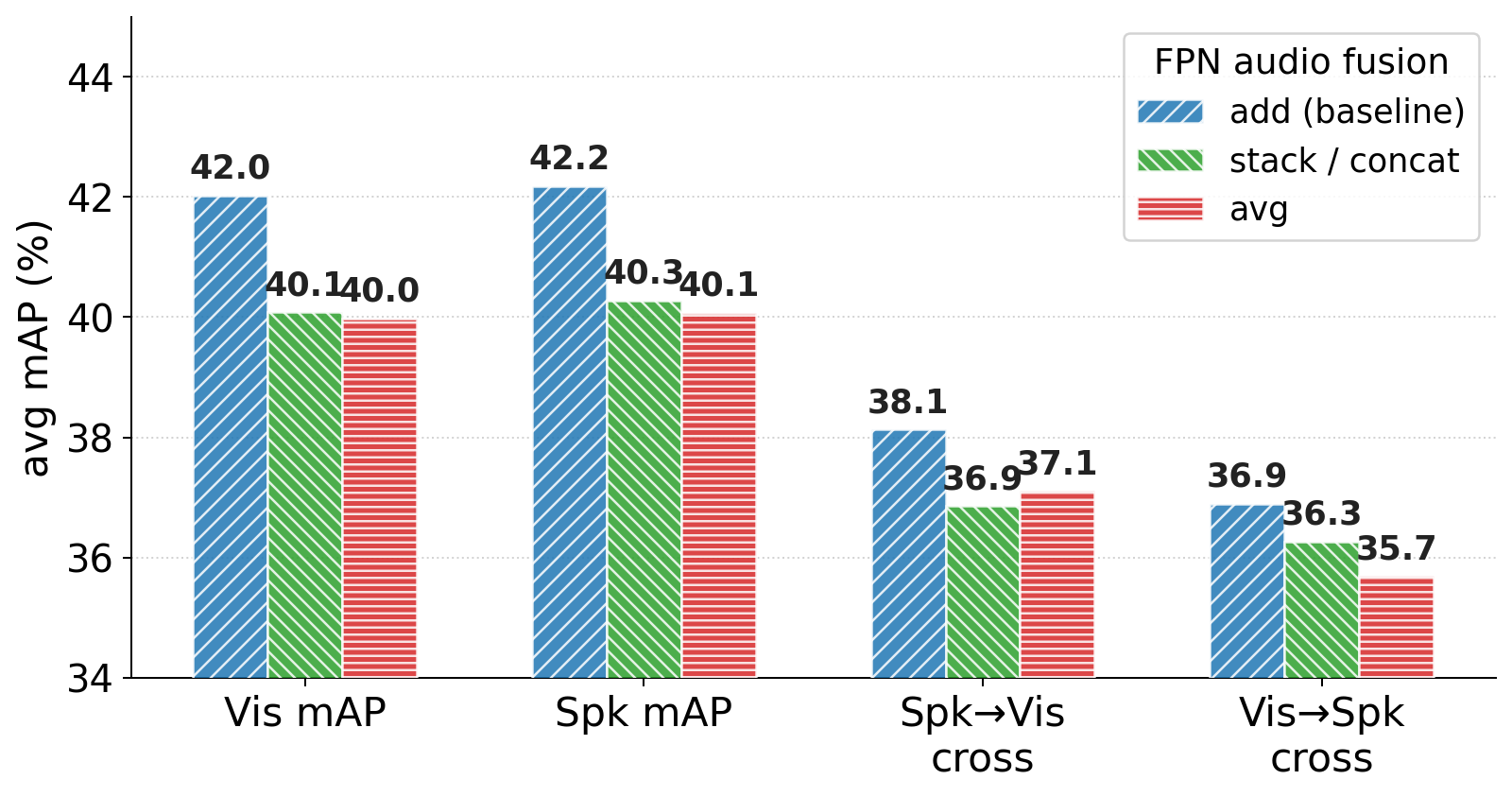}
    \caption{}
    \label{fig:fusion_ablation}
  \end{subfigure}
  \vspace{-0.5cm}
  \caption{\textbf{Stage~1 Dual-Head Analysis.}
  (\subref{fig:cross_head}) Each head evaluated against its own GT
  (\textbf{Same-head}) and against the other head's GT (\textbf{Cross-head}); the $\sim$4--5 point drop confirms the visual and spoken heads learn distinct temporal targets rather than a shared window.
  (\subref{fig:fusion_ablation}) Audio-fusion strategies (add, stack, avg) compared on visual mAP, spoken mAP, and the two cross-head settings. Element-wise addition consistently wins, justifying our fusion choice.}
    \label{fig:stage1_analysis}
\end{figure}

\begin{figure*}[t]
  \centering
  \begin{subfigure}{0.39\textwidth}
    \includegraphics[width=\linewidth]{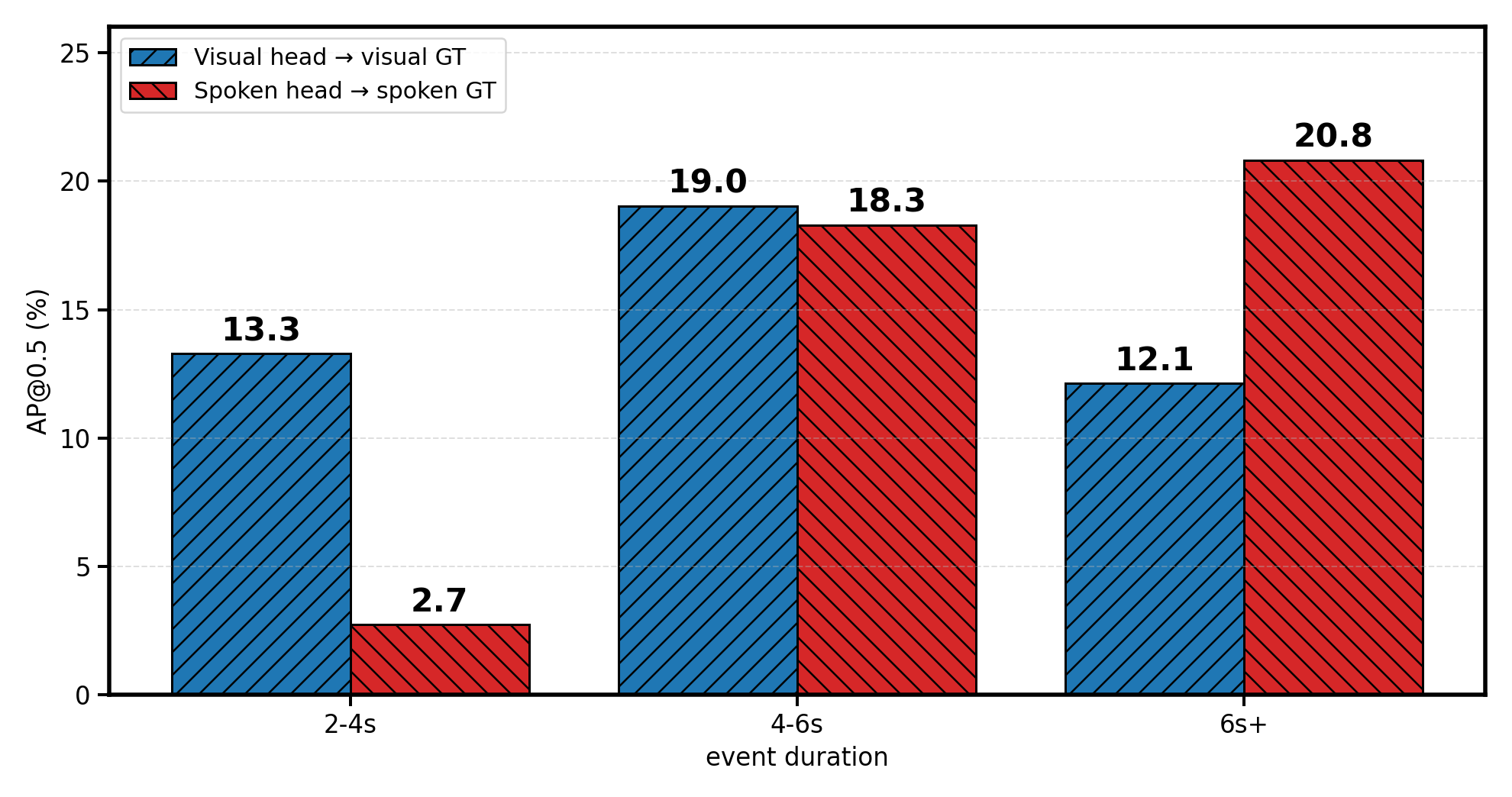}
    \caption{}
    \label{fig:dur_bins}
  \end{subfigure}
  \begin{subfigure}{0.59\textwidth}
    \includegraphics[width=0.95\linewidth]{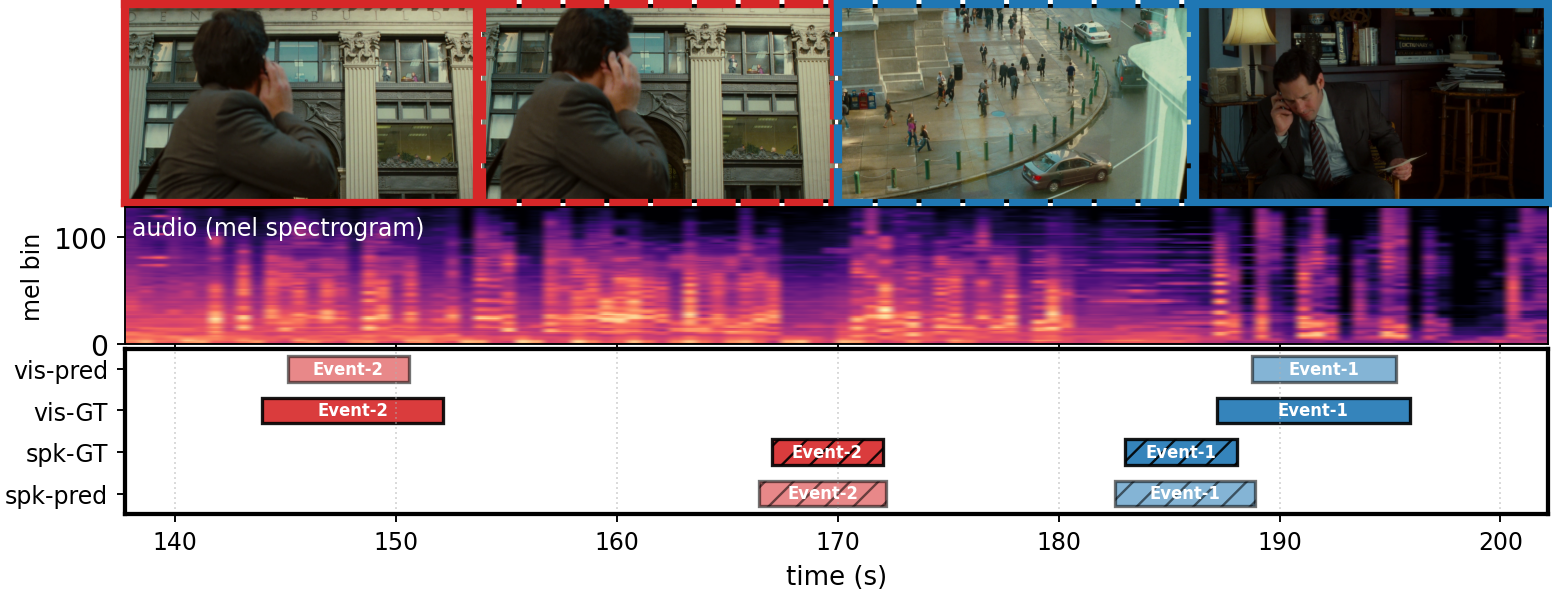}
    \caption{}
    \label{fig:dur_qual}
  \end{subfigure}
  \vspace{-0.1cm}
  \caption{\textbf{Modality Contribution.}
  (\subref{fig:dur_bins}) Per-instance AP@$0.5$ on paired annotations bucketed
  by event duration, for the visual head against the visual GT window (blue)
  and the spoken head against the spoken GT window (red).
  (\subref{fig:dur_qual}) Qualitative example with mel-spectrogram and
  frame strip with vis-pred/vis-GT and spk-pred/spk-GT tracks, illustrating
  the temporal offset between paired visual and spoken windows that
  \stageone recovers.}
    \label{fig:duration_qual}
\end{figure*}

\subsection{Localization Results}
\label{sec:loc_results}

Table~\ref{tab:stage1_mAP_by_test} reports localization on \longlsmdc with
zero-shot transfer to \cmdad and \madeval; we also report the fine-tuning results for the former; no additional pretraining data is
used. \stageone achieves 42.2 visual and 42.0 spoken mAP on
\longlsmdc, outperforming video-ActionFormer by $11.8$ points and the audio-only
variant by $5.0$ points. The gain over ActionFormer confirms that audio is a
critical modality: silence patterns directly signal narration slot availability,
information unavailable to a video-only model. The residual gap between
audio-only and the full model reflects the complementary role of video, which
identifies which silences correspond to narratively salient events. On \cmdad, \stageone reaches 41.0 spoken mAP (FT), above
\textit{Audio}-ActionFormer ($39.2$). On \madeval, it achieves 70.0
visual mAP, above ActionFormer ($63.7$). CA3D~\cite{Lee2025CA3D}, the only
prior AD localizer, reports $65.3$ F1 on \madeval; our model reaches
90.0 F1 on the same split (details in Appendix~\ref{sec:madeval_f1}), confirming the gain is not a metric
artifact. Furthermore, to have fair multi-modal baseline comparison, we extend ActionFormer with a dual-head architecture and naive audio-visual fusion which gives 38.5 average mAP. As a closed-source ablation, we probe Gemini~2.5~Pro~\cite{comanici2025gemini} for localization via prompted timestamp prediction; despite its strong video understanding, it
fails to produce reliable temporal boundaries, highlighting that precise AD
localization requires dedicated training (per-tIoU breakdowns in Appendix~\ref{sec:iou_analysis}).
All \ours results are means over three seeds, with standard deviation at most 0.2 mAP. The complete three-seed localization results are reported in Appendix Table~\ref{tab:multiseed_localization}.

\noindent\textbf{Dual-head design and fusion strategy.}
We ablate the two key design choices: The dual-head structure and how the
audio stream is fused into the visual FPN.
Fig.~\ref{fig:cross_head} shows the heads are not redundant: each is strong
on its own target but degrades by $4$--$5$ points when evaluated against the
other head's GT. The drop reflects the fact that paired visual and spoken windows in \longlsmdc are routinely offset in time
(Sec.~\ref{sec:dataset}); a single shared head would be forced to predict
one window or an average of the two, losing the offset that defines AD
timing. The dual-head design preserves it.
Fig.~\ref{fig:fusion_ablation} compares three strategies for combining the
audio stream with the visual FPN: addition (ours), stack/concat, and avg. Addition
wins consistently across all four settings, with concat and avg trailing by
$1.5$--$2$ mAP. Concat forces the model to learn a new joint representation
at each pyramid level, diluting the modality-specific signal; avg dampens both
streams symmetrically. Addition keeps each modality's contribution intact
and lets each head exploit them independently, audio for silence cues,
video for visual salience, which is exactly the complementarity the
dual-head design was set up to capture.

\noindent\textbf{Per-instance and qualitative evidence: each modality dominates its
regime.}
Audio reliably finds the dialogue gaps where narration can be inserted but cannot say which gaps are worth narrating; video supplies the complementary signal — shot changes, character actions, and on-screen events that distinguish a narratable moment from an arbitrary pause. 
Fig.~\ref{fig:duration_qual}a shows this
split at the instance level on paired annotations bucketed by event
duration. 
The spoken head (red) is weak at $2$--$4$\,s
($\text{AP@}0.5\!=\!2.7$) where dialogue gaps are too short to disambiguate
but dominates at $6$\,s+ ($20.8$) where the gap structure is unambiguous:
Its strength scales with the extent to which the gap is observable. The visual head (blue) is comparatively flat across durations ($13.3$ vs.\ $12.1$).
Since visual evidence accumulates over scene context rather than gap length, a short and a long event can carry a comparable visual signal. The two
modalities cover non-overlapping regimes. Fig.~\ref{fig:duration_qual}b
makes this concrete on a paired clip with two events: the mel-spectrogram
shows a dialogue dip in each, which the spoken head (\texttt{spk-pred})
aligns with, while the visual head (\texttt{vis-pred}) locks onto the
on-screen action visible in the frame strip. The \visual and \spoken for the same event are temporally offset; the action and its
narration do not coincide, and \stageone recovers both sides of the
offset.

\begin{table}[t]
\small
\centering
\caption{\textbf{Stage~2 generation on \longlsmdc and \cmdad.} Results use ground-truth \spoken segments; \longlsmdc is evaluated on \textbf{paired data} with \spoken, and \cmdad uses CharBank when available. Code is unavailable for AutoAD-III and DistinctAD, so only their reported \cmdad numbers are included. TF = training-free; C = CIDEr; AS = Action Score; R@1/5 = Recall@1/5. $^\dagger$UniAD is run on \longlsmdc from the released MAD-pretrained checkpoint.
}
\label{tab:stage2_generation}
\scriptsize
\setlength{\tabcolsep}{4pt}
\renewcommand{\arraystretch}{1.3}
\begin{tabular}{l c | ccc | ccc}
\Xhline{1pt}
\multirow{2}{*}{\textbf{Method}} 
& \multirow{2}{*}{\textbf{TF}}
& \multicolumn{3}{c|}{\cellcolor{lightgreen}\textbf{\longlsmdc}} 
& \multicolumn{3}{c}{\cellcolor{lightyellow}\textbf{\cmdad} (Zero-shot)} \\
& 
& \textbf{C} & \textbf{AS} & \textbf{R@1/5}
& \textbf{C} & \textbf{AS} & \textbf{R@1/5} \\
\Xhline{1pt}
AutoAD-III
& \xmark 
& -- & -- & --
& {25.0} & 31.5 & 31.2 \\
DistinctAD
& \xmark 
& -- & -- & --
& 22.7 & -- & 33.0 \\
UniAD$^\dagger$
& \xmark
& 10.1 & -- & --
& 21.8 & -- & -- \\
AutoAD-Zero
& \cmark
& 14.2 & -- & 31.3
& 17.7 & 25.5 & 26.9 \\
Shot-by-Shot 
& \cmark
& 16.9 & 25.3 & 28.4
& \textbf{26.3} & 28.4 & 33.0 \\
\midrule
Qwen2.5-VL-7B
& \xmark
& 30.1 & 41.7 & 43.8
& 20.5 & 35.4 & 34.0 \\
\quad + \textbf{\stagetwo}
& \xmark
& 32.5 & 40.9 & 43.3
& 22.3 & 35.0 & 33.2 \\
Qwen3-VL-8B
& \xmark
& 36.0 & 42.0 & 46.8
& 23.8 & 35.1 & 34.1 \\
\quad + \textbf{\stagetwo}
& \xmark
& \textbf{37.3} & \textbf{42.6} & \textbf{47.0}
& 25.1 & \textbf{36.0} & \textbf{34.4} \\
\Xhline{1pt}
\end{tabular}
\end{table}

\begin{table}[t]
\centering
\caption{\textbf{Stage~1+2 generation on LongLSMDC and CMD-AD.} Results are computed under Stage~1 predicted segments using tIoU@0.3 matching. Results on paired data. TF = training-free; C=CIDEr; AS=Action Score; R@1/5=Recall@1/5.}
\label{tab:stage1and2_generation}
\footnotesize
\setlength{\tabcolsep}{4pt}
\renewcommand{\arraystretch}{1.3}
\scalebox{0.85}{
\begin{tabular}{lc|ccc|ccc}
\Xhline{1pt}\noalign{\vskip 1pt}
\multirow{2}{*}{\textbf{Method}} &
\multirow{2}{*}{\textbf{TF}}
& \multicolumn{3}{c|}{\cellcolor{lightgreen}\textbf{\longlsmdc}} 
& \multicolumn{3}{c}{\cellcolor{lightyellow}\textbf{\cmdad (Zero-Shot)}} \\
& & \textbf{C} & \textbf{AS} & \textbf{R@1/5}
& \textbf{C} & \textbf{AS} & \textbf{R@1/5} \\
\noalign{\vskip 1pt}\Xhline{1pt}\noalign{\vskip 1pt}
AutoAD-Zero & \cmark
& 5.1 & 20.7 & 23.4
& 10.1 & \textbf{30.9} & 25.3 \\
\midrule
Qwen2.5-VL-7B & \xmark
& 21.7 & 38.8 & 37.3
& 7.5 & 23.4 & 24.8 \\
\quad + \textbf{\ours} & \xmark
& 22.2 & 38.3 & 37.0
& 9.2 & 25.1 & 24.9 \\
Qwen3-VL-8B & \xmark
& 23.8 & 39.4 & 39.3
& 8.8 & 24.8 & 25.2 \\
\quad + \textbf{\ours} & \xmark
& \textbf{25.2} & \textbf{40.2} & \textbf{40.1}
& \textbf{11.5} & 27.8 & \textbf{26.0} \\
\noalign{\vskip 1pt}\Xhline{1pt}
\end{tabular}}
\end{table}

\noindent\textbf{End-to-end coverage.}
Table~\ref{tab:coverage} evaluates predicted spoken windows using greedy
one-to-one matching. Precision is the fraction of predictions matched to
a GT \spoken, recall the fraction of GT spoken windows matched, and F1 (mean). Accordingly, unmatched predictions equal $1-P$ and missed
narration slots $1-R$.

\begin{table}[t]
\begingroup
\centering\small
\caption{\textbf{End-to-end spoken-window localization coverage on \longlsmdc validation.} We use greedy one-to-one matching at tIoU thresholds 0.3 and 0.5.}
\resizebox{\columnwidth}{!}{\begin{tabular}{lccccc}
\toprule
tIoU & Precision & Recall & F1 & Unmatched pred. & Missed slots\\
\midrule
0.3 & 27.5 & 94.7 & 42.7 & 72.5\% & 5.3\%\\
0.5 & 21.2 & 73.0 & 32.9 & 78.8\% & 27.0\%\\
\bottomrule
\end{tabular}}
\label{tab:coverage}
\endgroup
\end{table}

\subsection{Generation Results}
\label{sec:gen_results}

\begin{figure}[t]
\centering
\begin{minipage}[c]{0.52\columnwidth}
\centering
\scriptsize
\scalebox{0.87}{
\renewcommand{\arraystretch}{1.15}
\begin{tabular}{lcc}
\Xhline{0.8pt}
\textbf{Method} & \textbf{C} & \textbf{LLM} \\
\Xhline{0.8pt}
Gemini-2.5 Pro & 17.3 & 2.80 \\
Qwen2.5 + \stagetwo & 32.5 & 2.43 \\
Qwen3 + \stagetwo & \textbf{37.3} & 2.68 \\
\Xhline{0.8pt}
\end{tabular}}
\end{minipage}\hfill
\begin{minipage}[c]{0.48\columnwidth}
\centering
\includegraphics[width=\linewidth]{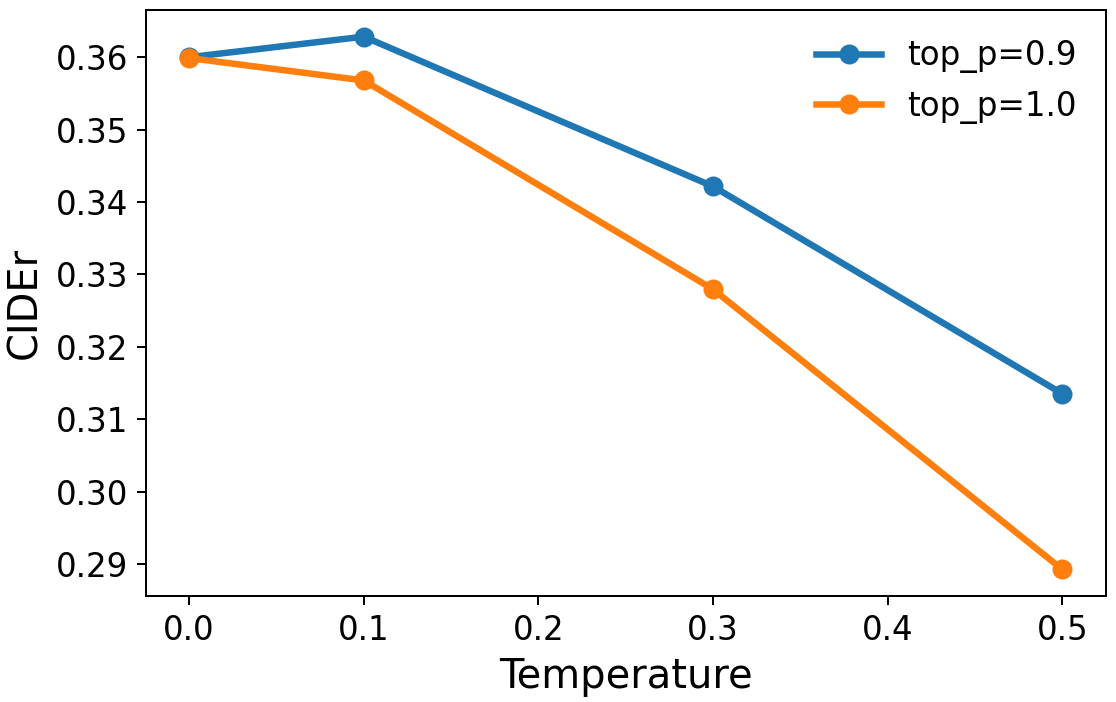}
\end{minipage}
\caption{\textbf{Stage~2: \longlsmdc spoken windows.}
\textbf{Left:} comparison against Gemini-2.5~Pro under GT$\rightarrow$S2.
\textbf{Right:} decoding-strategy ablation for \stagetwo; greedy decoding is is on par among settings tested. All Stage~2 results use greedy decoding.}
\label{fig:spoken_table_decode}
\end{figure}

We further extend to the \emph{what} component of AD, generating the
sentence that describes the event.

\noindent\textbf{Baseline convention.} In Tables~\ref{tab:stage1and2_generation} and~\ref{tab:stage2_generation}, the Qwen2.5-VL-7B and Qwen3-VL-8B rows report numbers \emph{after SFT on \longlsmdc AD annotations} (\ie the same LoRA adaptation, prompt, and training data as \stagetwo, but with $\mathcal{L}_{sft}$ only). The \texttt{+\,\stagetwo} rows add $\mathcal{L}_{drl}$ on top of this SFT baseline.
In all Stage~2 experiments, each AD is generated independently from its corresponding \visual; neither previously generated nor ground-truth neighbouring descriptions are provided as context.

\noindent\textbf{Generation under predicted and GT spoken windows.}
Tables~\ref{tab:stage1and2_generation} and~\ref{tab:stage2_generation}
report generation quality in two settings. We evaluate on paired spoken annotations.
\textbf{S1$\rightarrow$S2} (Table~\ref{tab:stage1and2_generation}) \stagetwo generates from \stageone predicted visual windows and is scored against GT ADs matched through the predicted spoken window (Sec.~\ref{sec:joint_eval}). \textbf{GT$\rightarrow$S2}
(Table~\ref{tab:stage2_generation}) replaces predicted windows with GT spoken segments, removing \stageone localization error from the evaluation and measuring generator quality in isolation.
In both settings, \stagetwo
improves over the base model on both backbones and both datasets: on
on \longlsmdc{}, $+1.3$ CIDEr for Qwen3-VL
($36.0\!\rightarrow\!37.3$) under GT windows and $+1.4$ CIDEr
($23.8\!\rightarrow\!25.2$) under predicted windows. The gain is
consistent across backbones, datasets, and evaluation settings, and
exceeds the seed-level standard deviation by roughly a factor of four
(Table~\ref{tab:multiseed_generation}), indicating that DRL adds a reliable
improvement on top of standard fine-tuning.
On \cmdad, the gain transfers zero-shot
($8.8\!\rightarrow\!11.5$ predicted; $23.8\!\rightarrow\!25.1$ GT).
Separate three-seed generation stability results are provided in Appendix Tab.~\ref{tab:multiseed_generation}.
For Qwen3-VL on \longlsmdc, CIDEr is 37.3 under GT windows and 25.2 under predicted windows. This difference should not be interpreted as a pure localization penalty because the predicted-window evaluation includes only successfully matched predictions, whereas the GT-window evaluation uses the complete paired GT set.
On \cmdad, the generator recovers when oracle input segments are provided ($25.1$ CIDEr) but drops sharply under predicted windows ($11.5$ CIDEr). Stage~1 still predicts a visual window for \cmdad, but \cmdad provides no GT visual annotations, so the generator was never exposed to \cmdad's distribution of visual windows; the predicted visual windows come from a localizer trained on \longlsmdc, while \cmdad clips are defined by narration-slot timing rather than the visual events. Thus \cmdad serves as a transfer benchmark. Its released MAD-pretrained checkpoint under GT spoken windows gives 10.1 CIDEr on \longlsmdc, against 34.3 for \ours in the matching no-CharBank setting (Table~\ref{tab:stage2_generation}). UniAD clearly outperforms the zero-shot Qwen3-VL backbone (5.3 CIDEr), confirming that the adaptation is faithful and that its contextual design transfers without target-domain training; the remaining gap reflects the distribution shift between UniAD's training data (short 2--3 min clips with sparse spoken-only slots) and \longlsmdc's dense, long-form clips with paired windows.

\noindent\textbf{Scope of the matched-pair evaluation.}
Text metrics in Table~\ref{tab:stage1and2_generation} (S1$\rightarrow$S2) are computed only on predictions that match a GT spoken window at tIoU~$>0.3$, since our proposed setting requires a visual--spoken pair to be present before an AD can be assessed against a reference. This mirrors how AD is produced in practice: an unmatched prediction has no reference sentence to score against. As a consequence, Table~\ref{tab:stage1and2_generation} reports quality \emph{conditional on successful localization} and does not penalize unmatched predictions or missed GT segments; localization coverage is reported separately in Table~\ref{tab:coverage}. Table~\ref{tab:stage2_generation} (GT$\rightarrow$S2) evaluates on the complete paired GT set and is directly comparable to prior SOTA numbers on \cmdad. Because the two settings use different reference pools, their CIDEr values should not be interpreted as a direct estimate of the localization penalty.

\begin{figure}[t]
    \begin{tikzpicture}
        \node[anchor=south west, inner sep=0] (img2) at (0,0)
            {\includegraphics[width=\linewidth]{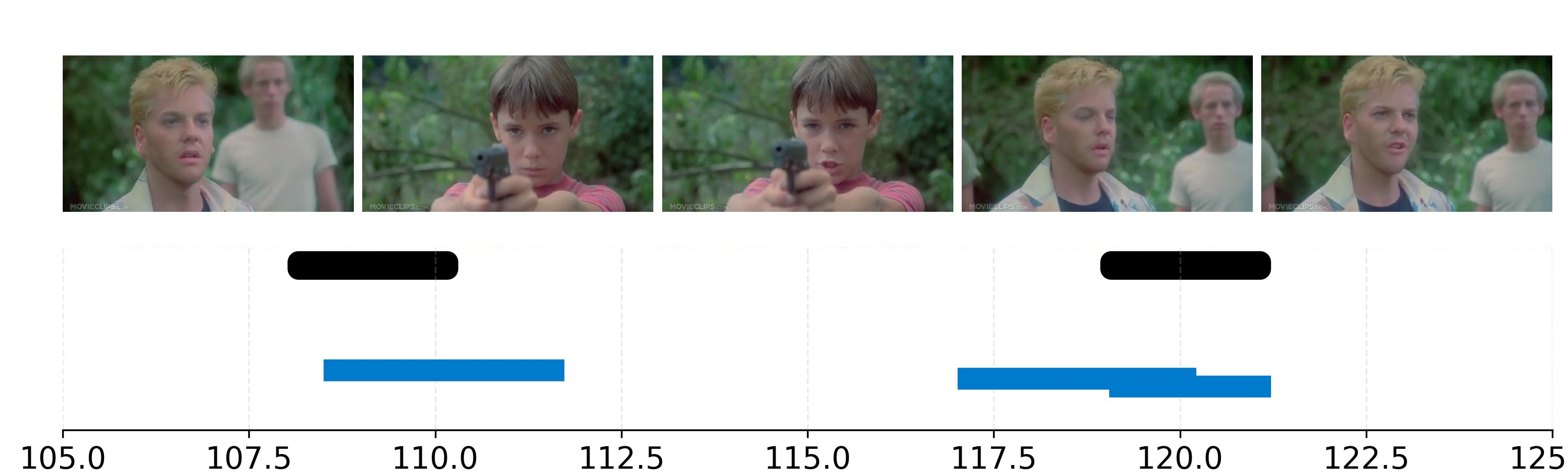}};
        \begin{scope}[x={(img2.south east)}, y={(img2.north west)}]
            \node[anchor=north west, font=\tiny, text=white, fill=red, fill opacity=0.2, text opacity=1, inner sep=0pt] at (0.195,0.495) {GT 1};
            \node[anchor=north west, font=\tiny, text=white, fill=red, fill opacity=0.2, text opacity=1, inner sep=0pt] at (0.716,0.495) {GT 2};
            \node[anchor=north west, font=\tiny, text=white, fill=gray, fill opacity=0.6, text opacity=1, inner sep=0pt] at (0.23,0.275) {Pred 1};
            \node[anchor=north west, font=\tiny, text=white, fill=gray, fill opacity=0.6, text opacity=1, inner sep=0pt] at (0.64,0.260) {Pred 2};
            \node[anchor=north west, font=\tiny, text=white, fill=gray, fill opacity=0.6, text opacity=1, inner sep=0pt] at (0.73,0.240) {Pred 3};
        \end{scope}
    \end{tikzpicture}\\[1mm]
    \noindent\makebox[\linewidth][l]{\footnotesize GT 1: Gordy pulls the hammer.}
    \noindent\makebox[\linewidth][l]{\footnotesize Pred 1: Ace stands and watches Chris walk away in the woods.}
    \noindent\makebox[\linewidth][l]{\footnotesize GT 2: Ace swallows and glances at Charlie.}
    \noindent\makebox[\linewidth][l]{\footnotesize Pred 3: Ace lowers his gaze and shifts his weight.}
    \caption{\textbf{Qualitative example on \cmdad.} Top: sampled frames.
    Bottom: Pred segments (blue) vs.\ GT (black). GT/Pred AD
    sentences are listed below.}
    \label{fig:results1}
\end{figure}

\noindent\textbf{Qualitative analysis on dense, long-form clips.}
Fig.~\ref{fig:results1} shows a \cmdad clip with two AD events: \stagetwo
recovers both spoken windows and generates contextually relevant
descriptions, though phrasing diverges from the reference (e.g.,
\textit{``Ace stands and watches Chris walk away''} vs.\ GT
\textit{``Gordy pulls the hammer''}) a common pattern when multiple
plausible visual subjects share the frame. Dense, multi-event examples
from \longlsmdc are in Appendix~\ref{sec:qual_both}.
This substitution pattern is one of four failure modes we categorise in Appendix~\ref{sec:qual_both}: (i) wrong-subject attribution when multiple plausible visual subjects share the frame, (ii) correct event but an under-specified verb, (iii) boundary drift producing a description of adjacent content, and (iv) hallucinated detail. Representative examples are given there.

\noindent\textbf{Comparison against closed-source models and decoding ablation.}
Fig.~\ref{fig:spoken_table_decode} probes two questions about \stagetwo.
\textbf{Left:} how does \stagetwo compare to a strong closed-source VLM? Gemini-2.5~Pro~\cite{comanici2025gemini}, evaluated in the \textbf{GT$\rightarrow$S2} setting, trails both \stagetwo backbones on CIDEr by a wide margin ($17.3$ vs.\ $32.5$ / $37.3$) while scoring highest on LLM-Eval ($2.80$ vs.\ $2.43$ / $2.68$). The two metrics measure different properties: LLM-Eval rewards coherent, fluent narration, whereas CIDEr rewards $n$-gram overlap with the concise reference-AD style. Gemini produces fluent free-form descriptions that a judge model rates well, but does not conform to the compact AD register that Cue2Narrate's Description Ranking Loss explicitly trains for, which the CIDEr gap reflects. Tables~\ref{tab:stage1and2_generation} and~\ref{tab:stage2_generation} isolate the DRL contribution: the \texttt{+\,\stagetwo} rows, which add $\mathcal{L}_{drl}$ on top of the SFT baseline described above, consistently improve CIDEr, showing that DRL provides gains beyond standard fine-tuning.
\textbf{Right:} how sensitive is \stagetwo to decoding choices? Greedy decoding gives the best result,
indicating \stagetwo's output
distribution is sharp enough that sampling rarely finds better phrasings.

\section{Conclusion}
\label{sec:conclusion}

We re-framed automatic Audio Description as the joint prediction of \emph{what} to narrate and \emph{when} to narrate it, and introduced \longlsmdc to benchmark this problem on long-form movie clips. \ours pairs a dual-head audio-visual localizer with a window-conditioned generator. The dual head separately predicts where the event is visible and where its narration is voiced, recovering the offset between the two that defines AD timing. 
We hope \longlsmdc will facilitate new exciting research in automatic AD generation.

\section{Limitations}
\label{sec:limitations}

This work is a first step toward automatic audio description on long-form, untrimmed video, and opens up several directions. (i) Two-stage training. Our pipeline keeps localization and generation separate; jointly training the two stages would let generation feedback shape window prediction and is a natural next step. (ii) Cross-description context. Stage~2 conditions on the visual cue window and generates each AD independently. Extending the generator to reason over the spoken narration window and adjacent ADs (surrounding dialogue, previously voiced descriptions) could further improve coherence across consecutive descriptions. (iii) User study. A study with blind and low-vision users would complement our 20-evaluator pairwise study of coverage, timing, and narrative quality, and we leave it to future work as it is beyond the scope of this paper. (iv) Character banks. Character identity is assigned automatically; Appendix~\ref{sec:dataset_prep} quantifies the pipeline's agreement rate, and human verification of the banks is a natural extension.

\section{Ethical Considerations}
\label{sec:ethical}

\ours is intended as an assistive tool to expand AD coverage for blind and low-vision audiences, complementing rather than replacing human describers. Automatically generated ADs can be contextually plausible but can be factually wrong, which is particularly consequential for users who depend on them as their primary access to visual content; deployment should disclose machine generation and support human review. Both the training data and the underlying VLM carry demographic biases that may surface in generated descriptions, and auditing for descriptive fairness is an important direction we leave to future work. \longlsmdc annotations follow LSMDC's research-only terms; the underlying video content is not redistributed. Released annotations, clips and character banks inherit LSMDC's restricted-access research-only agreement and are distributed through the same request process; code, checkpoints and evaluation scripts are released without restriction. Character banks contain names of fictional characters only and no personal data about real individuals beyond what is already public in the source cast lists.

\section{Acknowledgment}
\label{sec:ack}

The research at TU Darmstadt was partially funded by a LOEWE-Spitzen-Professur (LOEWE/4a//519/05.00.002(0010)/93), an Alexander von Humboldt Professorship in Multimodal Reliable AI, sponsored by Germany’s Federal Ministry of Research, Technology, and Space (BMFTR) and has benefited from the Excellence
Cluster “Reasonable AI” by the German Research
Foundation (Deutsche Forschungsgemeinschaft - DFG) under Germany’s Excellence Strategy (EXC-3057). For compute, we gratefully acknowledge support from the hessian.AI Service Center (funded by the BMFTR, grant no. 16IS22091) and the hessian.AI Innovation Lab (funded by the Hessian Ministry for Digital Strategy and Innovation, grant no. S-DIW04/0013/003).

\bibliography{custom}

\appendix
\clearpage

In this supplementary material, we provide: (A) dataset, character-bank, and character-name agreement details (Sec.~\ref{sec:dataset_prep}); (B) fine-grained IoU evaluation, coverage metrics, the CA3D comparison, and the UniAD adaptation protocol (Sec.~\ref{sec:iou_analysis}); (C) human evaluation (Sec.~\ref{sec:human_eval}); (D) negative-caption examples (Sec.~\ref{sec:qwen_drl}); (E) Action-Score reproducibility and Gemini prompts (Sec.~\ref{sec:gemini_prompts}); (F) Stage~2 fine-tuning prompts (Sec.~\ref{sec:loc2narrate_prompts}); (G) additional qualitative results and systematic error categories (Sec.~\ref{sec:qual_both}); (H) artifact access and licenses (Sec.~\ref{sec:artifact}); and (I) model size and budget (Sec.~\ref{sec:model_size}).

\section{Additional \longlsmdc and Character-bank details}
\label{sec:dataset_prep}

The original LSMDC~\cite{Rohrbach2017LSMDC} consists of full-length movies (90--180 minutes)
with sparsely distributed AD annotations. Processing full-length movies end-to-end is
computationally prohibitive for both training and evaluation: a single 90-minute film
contains thousands of frames and hundreds of AD slots, making batched temporal localization
infeasible at movie scale. We therefore segment each movie into variable-length clips
(5--8 minutes), each containing multiple AD annotations, enabling scalable training and
dense multi-segment evaluation. 
\paragraph{Release plan.} We release: (i) all \longlsmdc annotations---paired visual and spoken windows, clip boundaries, and the train/validation splits; (ii) the preprocessed movie clips and extracted features, under the same terms as LSMDC and in coordination with the original authors; (iii) the per-clip character banks; (iv) the full training and inference code for both stages; (v) the trained Stage~1 localizer and Stage~2 LoRA adapters; and (vi) all evaluation scripts, including the matching protocol and the Action-Score pipeline. Items (i)--(iii) inherit LSMDC's restricted-access, research-only agreement and require the standard LSMDC request; items (iv)--(vi) are unrestricted. Underlying video is not redistributed. Full details and license terms are in Appendix~\ref{sec:artifact}.

\paragraph{Random clip segmentation.}
For each movie of duration $D$, we divide it into non-overlapping clips with randomized durations
$\delta \sim \mathcal{U}(300\text{s},\, 480\text{s})$.
Starting from $t = 0$, we create clip $[t,\, \min(t + \delta,\, D))$, then advance $t$ to the
clip end, continuing until the entire movie is segmented. Random lengths prevent
duration-specific biases and create variable clip complexity (1--147 annotations per clip).
The few clips shorter than 300\,s ($<$0.2\%) are tail segments produced when a movie ends
before the sampled duration expires.

\begin{figure}[t]
  \centering
  \includegraphics[width=\linewidth]{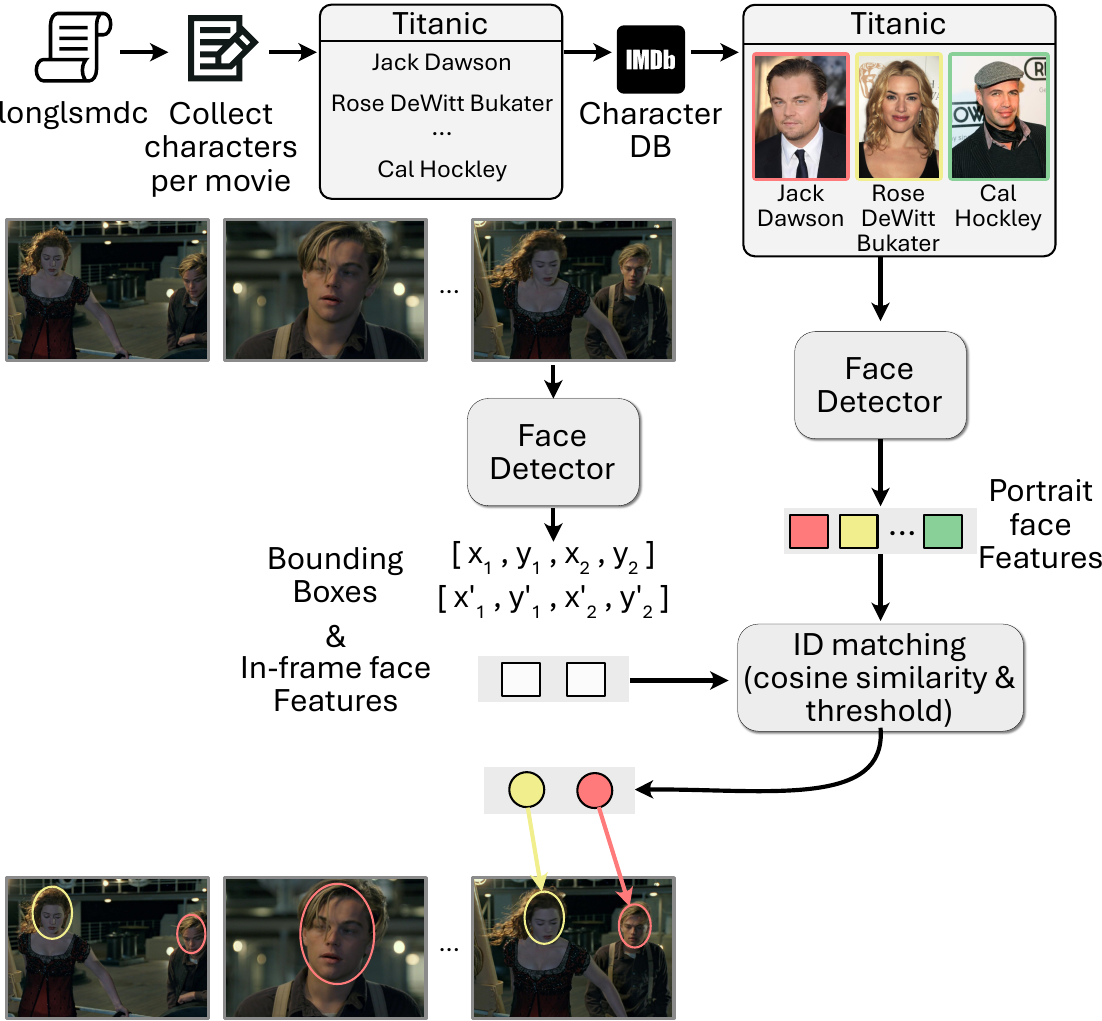}
  \caption{    \textbf{LongLSMDC character bank construction pipeline.}
    For each movie, we collect a cast list of character names from an external database.
    A face detector is applied to movie frames to extract bounding boxes and in-frame face
    features, which are compared against portrait-level face features for each character via
    cosine similarity with a fixed threshold, assigning detections to named characters.
    The resulting per-movie character bank maps frame-level detections to character identities
    (e.g., Jack Dawson, Rose DeWitt Bukater, and Cal Hockley in \textit{Titanic}),
    enabling CharBank-conditioned AD generation at inference time.
  }
  \label{fig:charbank}
\end{figure}

\begin{figure}[ht]
    \centering
    \begin{subfigure}[b]{0.48\textwidth}
        \centering
        \includegraphics[width=\textwidth]{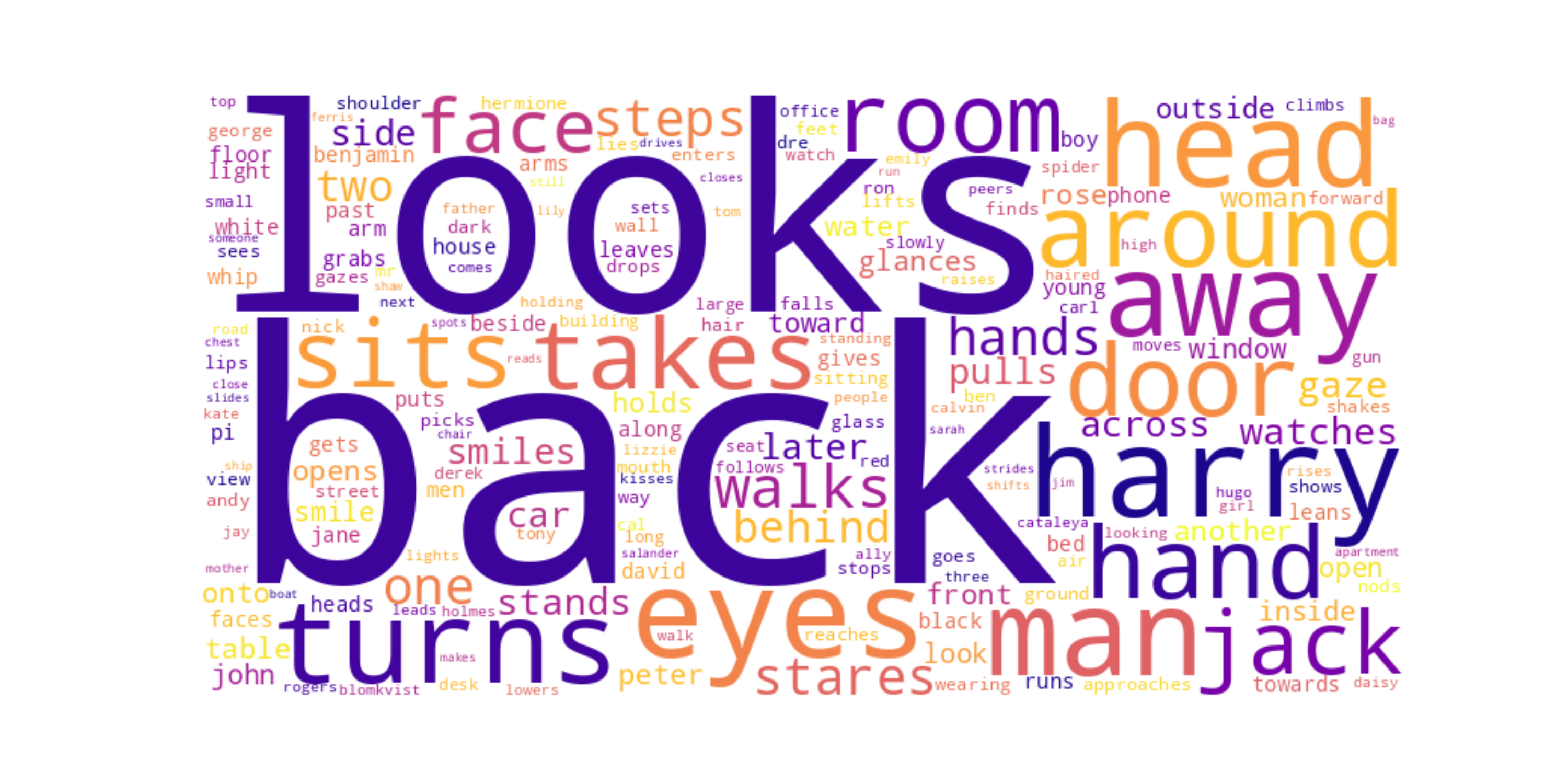}
    \end{subfigure}
    \hfill
    \begin{subfigure}[b]{0.48\textwidth}
        \centering
        \includegraphics[width=\textwidth]{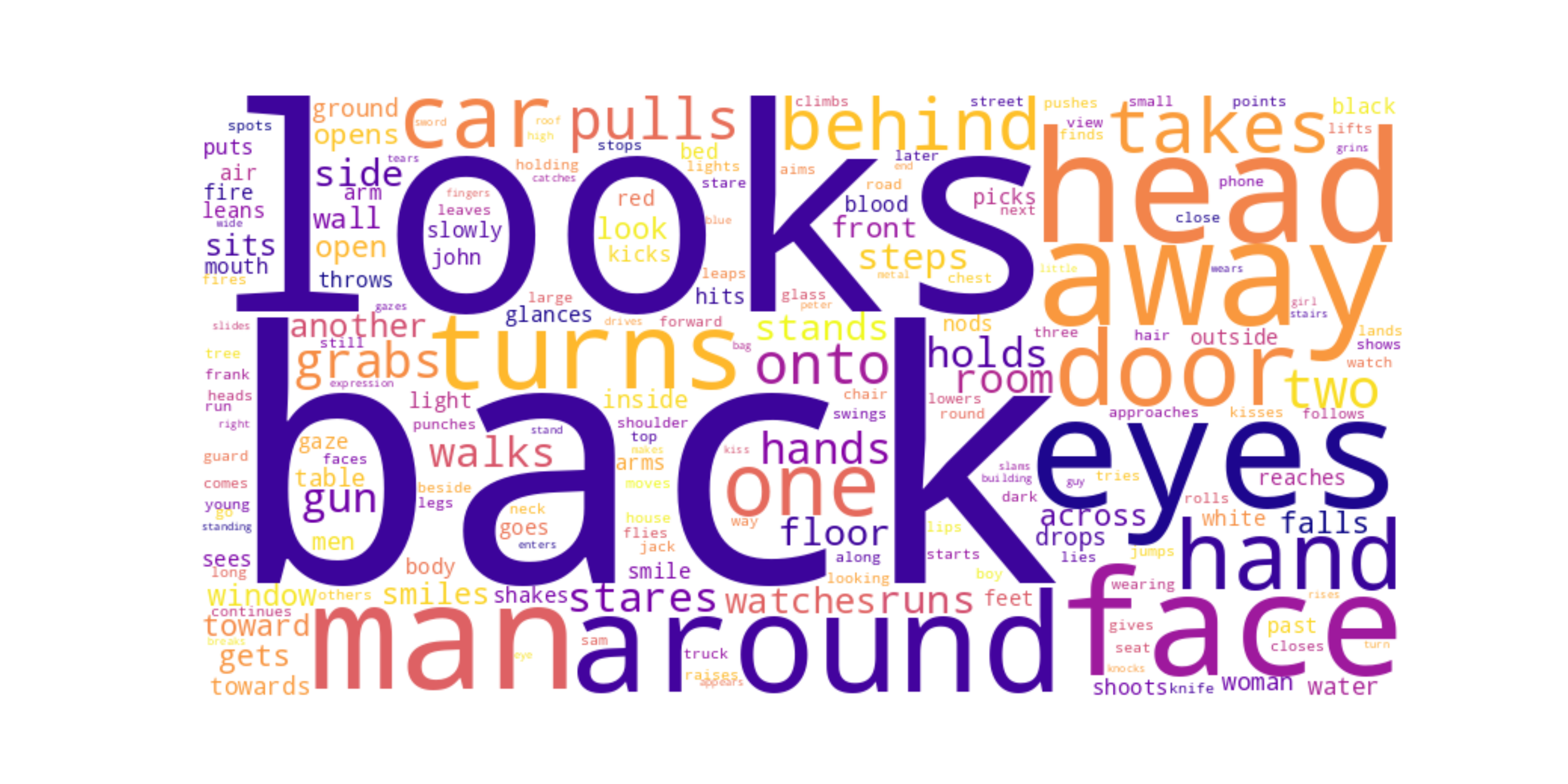}
    \end{subfigure}
    \vspace{-1mm}
    \caption{\textbf{Vocabulary comparison between \longlsmdc (Top) and \cmdad (Bottom) datasets.}
    \longlsmdc exhibits richer action verb diversity (e.g., ``walks'', ``turns'', ``looks'',
    ``stands'') and more varied character descriptors, reflecting its longer clips and broader
    movie coverage across diverse genres and cinematic styles.}
    \label{fig:visual_ad}
\end{figure}

\paragraph{Annotation assignment via midpoint rule.}
Each AD annotation $[s, e]$ is assigned to clip $c = [t_s, t_e)$ if its temporal midpoint
$m = (s + e)/2$ satisfies $t_s \leq m < t_e$. This ensures each annotation belongs to exactly
one clip and is temporally centered for maximum context. We compute clip-relative coordinates
$s' = s - t_s$ and $e' = e - t_s$ for standardized temporal ranges. Annotations on clip
boundaries ($<$1\%) are excluded.

\begin{figure*}[t]
    \centering
    \scalebox{0.98}{
    \begin{minipage}{0.33\textwidth}
        \centering
        \includegraphics[width=\linewidth]{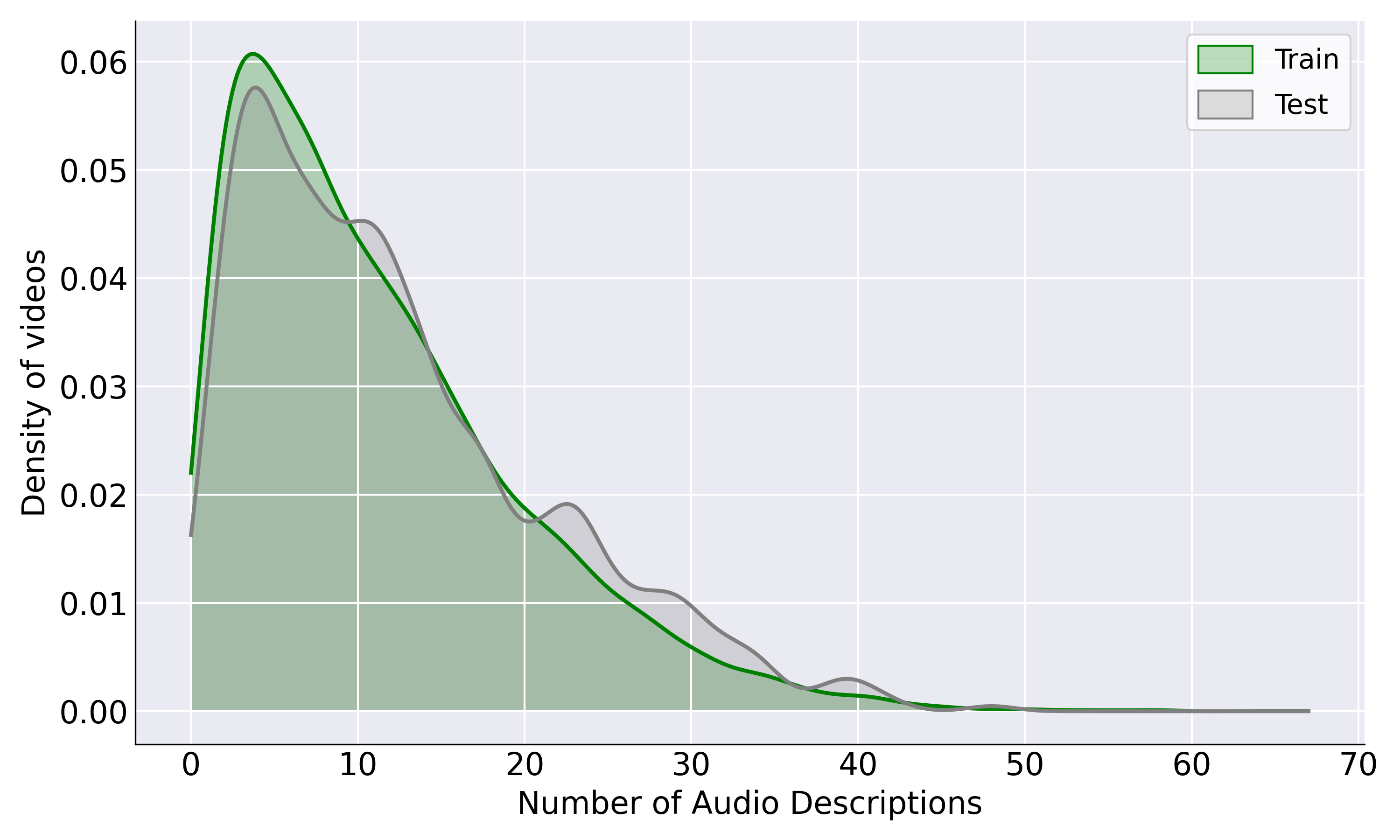}
        \label{fig:subfig3}
    \end{minipage}
    \begin{minipage}{0.33\textwidth}
        \centering
        \includegraphics[width=\linewidth]{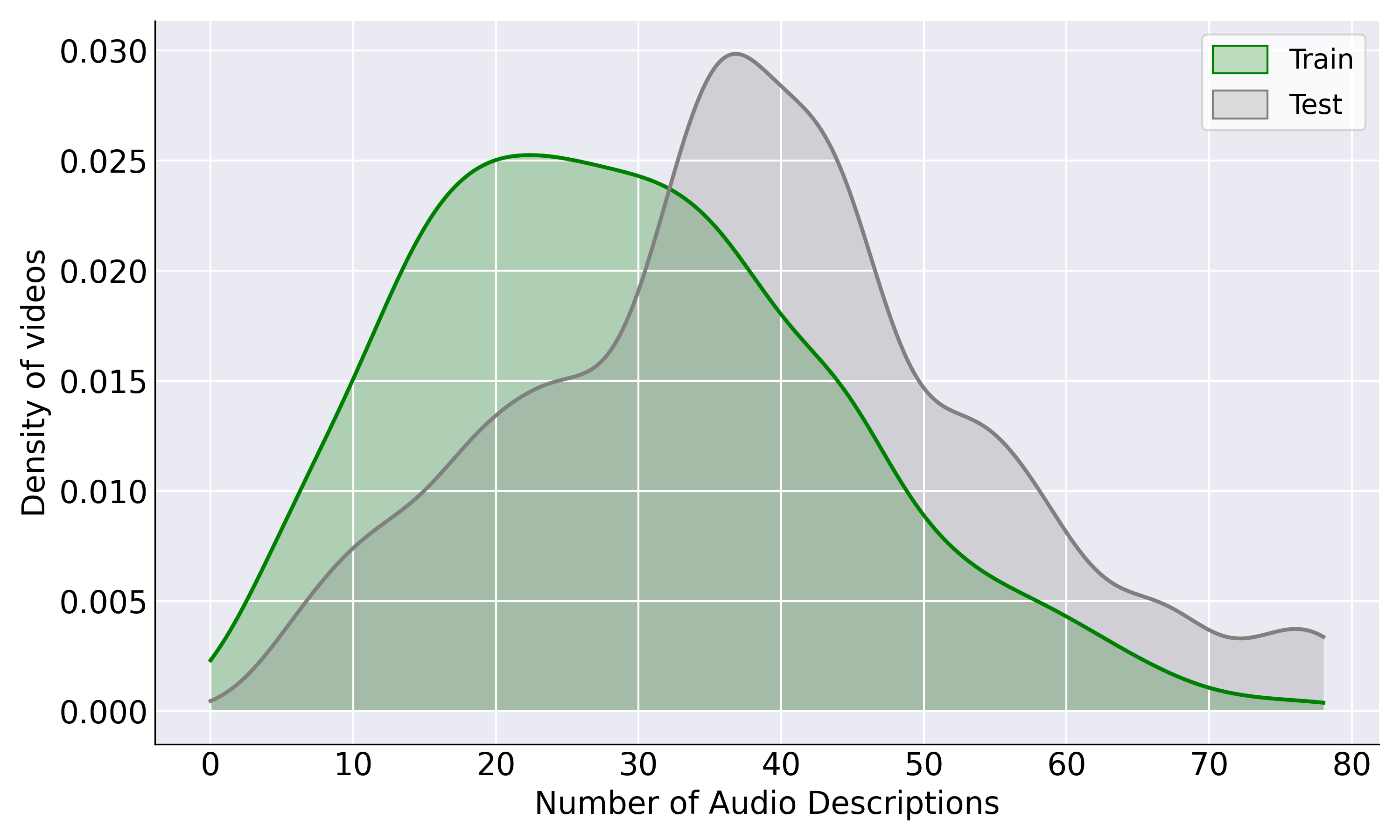}
        \label{fig:subfig2}
    \end{minipage}
    \begin{minipage}{0.33\textwidth}
        \centering
        \includegraphics[width=\linewidth]{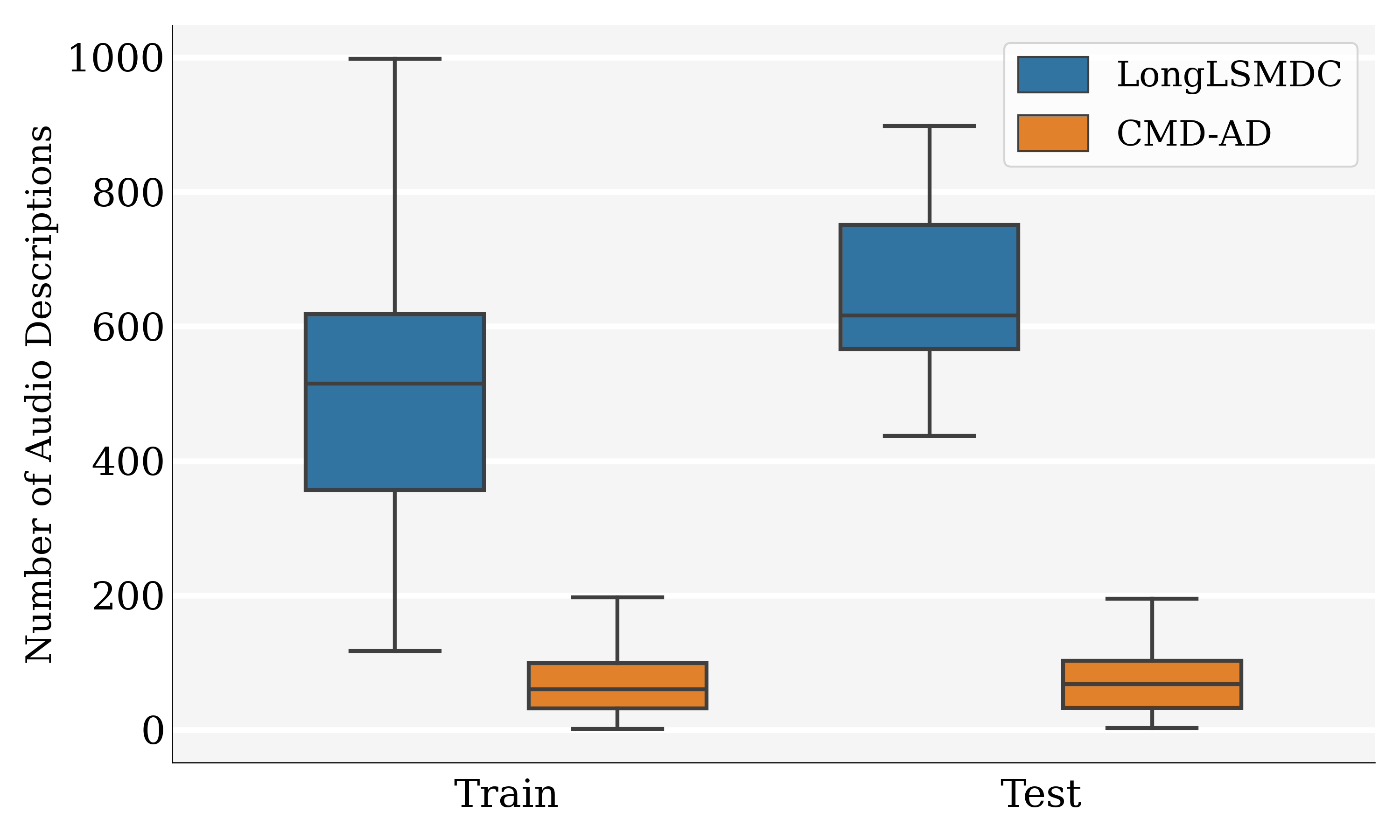}
        \label{fig:subfig1}
    \end{minipage}
    }
    \vspace{-4mm}
    \caption{Dataset statistics for \cmdad and \longlsmdc. (a) \cmdad contains short clips
    (2--3 min) with 1--5 audio descriptions per clip. (b) \longlsmdc contains longer clips
    (up to 8 min) with a much higher density of AD segments (29--40 per clip). (c) Boxplots of
    the full movie-level distributions show that \longlsmdc provides substantially more AD
    events per movie than \cmdad, supporting evaluation of long-form, multi-segment AD
    generation.}
    \label{fig:main}
\end{figure*}

This transforms $\sim$125 movies into 2{,}100$+$ clips (1{,}995 training, 170 validation)
with over 78{,}000 training annotations across all annotation categories. Individual clips
contain between 1 and 147 AD annotations (mean: 29--40 ADs per clip depending on annotation
category), providing diverse density conditions for localization evaluation. Dataset density
statistics comparing \longlsmdc and \cmdad are shown in Figure~\ref{fig:main}: \cmdad contains
short clips (2--3 min) with 1--5 ADs per clip, while \longlsmdc contains longer clips ( up to 8 min)
with 29--40 ADs per clip, providing substantially more AD events per movie and better supporting
evaluation of long-form, multi-segment AD generation.

\paragraph{Character bank construction.}
Following AutoAD-Zero \cite{Xie2024AutoADZero}, we construct a per-clip character bank
for \longlsmdc to support CharBank-conditioned generation experiments
(Figure~\ref{fig:charbank}). For each movie, we collect a cast list of character names from
IMDb. A face detector is applied to movie frames to extract bounding boxes and in-frame face
features, which are then compared against portrait-level face features for each character via
cosine similarity with a fixed threshold, assigning frame-level detections to named characters.
The resulting per-movie character bank maps detected faces to character identities (e.g.,
Jack Dawson, Rose DeWitt Bukater, and Cal Hockley in \textit{Titanic}), enabling
CharBank-conditioned AD generation at inference time. Unlike AutoAD-Zero, which constructs
character banks over ground-truth segments, our banks are built over full movie frames and
are therefore compatible with predicted windows. Character metadata is provided per clip for
both training and validation splits and will be released alongside the dataset.

\paragraph{Character-bank agreement analysis.} To measure how often the automatic pipeline finds the character an AD actually talks about, we check, for each character named in a GT AD, whether the bank identifies that character within the AD's window. Of 4,518 character mentions across validation ADs, 748 refer to characters with no portrait in the bank, so no match is possible for them by construction. Among the remaining 3,770 mentions, the bank identifies the character within the window in 2,655 cases (70.4\%). We manually inspected a sample of the non-identified cases and found the most frequent cause is not a matching failure at all: the narration references a character whose face is not visible in the window (off-screen, turned away, or shot from behind). No face-matching pipeline, however robust, resolves such cases. For the cases where matching is possible but visually difficult---extreme profiles, occlusion, low lighting---the end-to-end comparison in Sec.~\ref{sec:gen_results} (+3.0 CIDEr with CharBank, with consistent gains in Action Score and R@1/5) shows that residual matching errors do not propagate harmfully into generation.

\paragraph{Vocabulary analysis.}
Figure~\ref{fig:visual_ad} presents word cloud visualizations comparing \longlsmdc and \cmdad
vocabularies. \longlsmdc exhibits richer action verb diversity (e.g., ``walks'', ``turns'',
``looks'', ``stands'') and more varied character descriptors, reflecting its longer clips and
broader movie coverage across diverse genres and cinematic styles. This vocabulary richness
supports more comprehensive evaluation of generation models' ability to produce varied, natural
language descriptions across different narrative contexts.

\paragraph{Metrics.} 
We follow standard Temporal Action Localization (TAL) evaluation protocols~\cite{Tang2023TemporalMaxer,Zhang2022ActionFormer}, reporting mean Average Precision (mAP) at temporal IoU thresholds $tIoU \in \{0.2, 0.3, 0.4, 0.5, 0.6\}$. These thresholds are widely used to evaluate TAL methods and assess localization quality at varying precision levels. The temporal IoU (tIoU) is defined as the intersection over union between two temporal windows, equivalent to the 1D Jaccard index. \longlsmdc supplies two ground-truth views per opportunity, we report mAP for each head against each view, giving a $2\!\times\!2$ matrix that reveals whether the visual and spoken heads predict the same window or genuinely distinct targets

A predicted segment $(t_s^{pred}, t_e^{pred})$ matches ground-truth $(t_s^{gt}, t_e^{gt})$ if their temporal IoU exceeds the threshold:
\begin{equation}
  \text{IoU} = \frac{|[t_s^{pred}, t_e^{pred}] \cap [t_s^{gt}, t_e^{gt}]|}{|[t_s^{pred}, t_e^{pred}] \cup [t_s^{gt}, t_e^{gt}]|} \geq \text{threshold}
\end{equation}

For each ground-truth segment, we follow the greedy matching approach: we find the highest-IoU predicted segment that exceeds the threshold. Matched predictions are considered true positives (TP), unmatched predictions are false positives (FP), and unmatched ground-truth segments are false negatives (FN). Average Precision (AP) is computed per video using the precision-recall curve, and mAP is the mean across all test videos. Given a tIoU threshold, mAP computes the mean of average precision across all videos. 
\paragraph{Coverage metrics.} Alongside mAP we report the precision, recall and F1 of the predicted spoken narration windows against GT spoken windows under the same greedy one-to-one tIoU matching used for the text metrics. The unmatched-prediction rate is $1-P$ (predictions with no GT match, i.e., potential false insertions) and the missed-slot rate is $1-R$ (GT narration slots with no matched prediction). We report both the matching threshold used in the main paper (0.3) and a stricter threshold (0.5); see Table~\ref{tab:coverage}.

\paragraph{Coverage interpretation.} At tIoU 0.3, recall is 94.7\%, leaving 5.3\% of GT narration slots unmatched. At the stricter tIoU 0.5 threshold, recall falls to 73.0\%, reflecting increased sensitivity to boundary precision. The unmatched-prediction rate can overstate false insertions because Soft-NMS retains overlapping candidates and because \longlsmdc annotates the slots selected by a professional describer rather than every potentially usable dialogue gap. The operating point can therefore be adjusted when insertion-rate control is required.

For \emph{generation} we use four complementary metrics: \textbf{CIDEr}~\cite{vedantam2015cider} for n-gram consensus with reference captions, \textbf{Recall@$k$} ($k\!=\!1,5$) for embedding-space similarity via retrieval, \textbf{Action Score}~\cite{Xie2025ShotByShot} for verb correctness (the most informative AD attribute for the listener), and \textbf{LLM-Eval} (LLaMA3-8B~\cite{dubey2024llama} as judge, 1--5 score, mean reported).

\section{Fine-Grained IoU Analysis for \longlsmdc}
\label{sec:iou_analysis}

\begin{table}[t]
\begingroup
\centering
\small
\caption{\textbf{Multi-seed localization stability on \longlsmdc.} Mean $\pm$ standard deviation over three seeds; mAP is averaged over tIoU thresholds 0.2--0.6.}
\label{tab:multiseed_localization}
\begin{tabular}{lccc}
\toprule
Method & Visual mAP & Spoken mAP & Avg. mAP\\
\midrule
\ours & $42.2\!\pm\!0.1$ & $42.0\!\pm\!0.2$ & $42.1\!\pm\!0.1$\\
\bottomrule
\end{tabular}
\endgroup
\end{table}

\begin{table}[t]
\begingroup
\centering
\small
\caption{\textbf{Multi-seed generation stability on \longlsmdc.} Qwen3-VL-8B CIDEr, reported as mean $\pm$ standard deviation over three seeds.}
\label{tab:multiseed_generation}
\begin{tabular}{lcc}
\toprule
Evaluation & SFT & $+\,\mathcal{L}_{drl}$ (\ours)\\
\midrule
GT$\rightarrow$S2 & $36.0\!\pm\!0.3$ & $37.3\!\pm\!0.2$\\
S1$\rightarrow$S2 & $23.8\!\pm\!0.2$ & $25.2\!\pm\!0.1$\\
\bottomrule
\end{tabular}
\endgroup
\end{table}

\begin{figure*}[t]
    \centering
    \includegraphics[width=\linewidth]{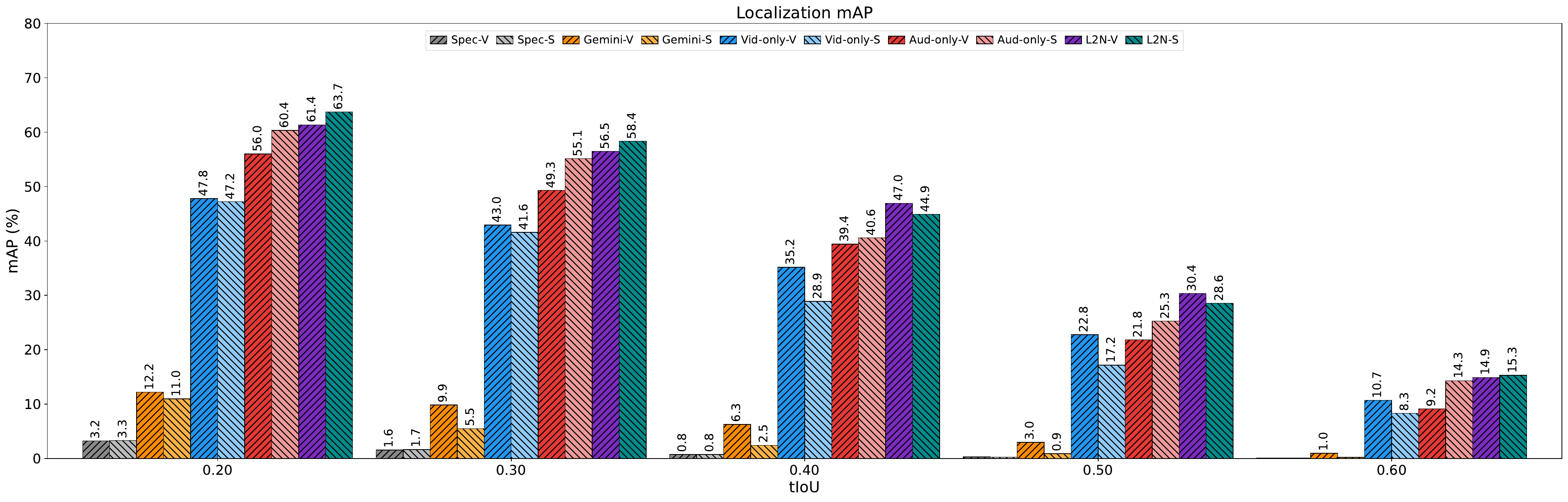}
    \caption{\textbf{Per-tIoU localization on \longlsmdc.}
    Grouped bar plot of mAP at tIoU thresholds $\{0.2,\,0.3,\,0.4,\,0.5,\,0.6\}$
    (x-axis), with bar height encoding mAP (\%). For each threshold, ten bars
    compare five methods evaluated against both the \visual head (\textbf{-V},
    hatched) and the \spoken head (\textbf{-S}, solid): Spectrogram (Spec),
    Gemini~2.5~Pro, video-only ActionFormer (Vid-only), audio-only
    ActionFormer (Aud-only), and \stageone (C2N). Gemini is competitive at
    loose tIoU but collapses at stricter thresholds; \stageone leads across
    all thresholds, with the margin over unimodal baselines widening as tIoU
    tightens.}
    \label{fig:per_iou_ablation}
\end{figure*}

\begin{figure*}[t]
    \centering
    \includegraphics[width=\linewidth]{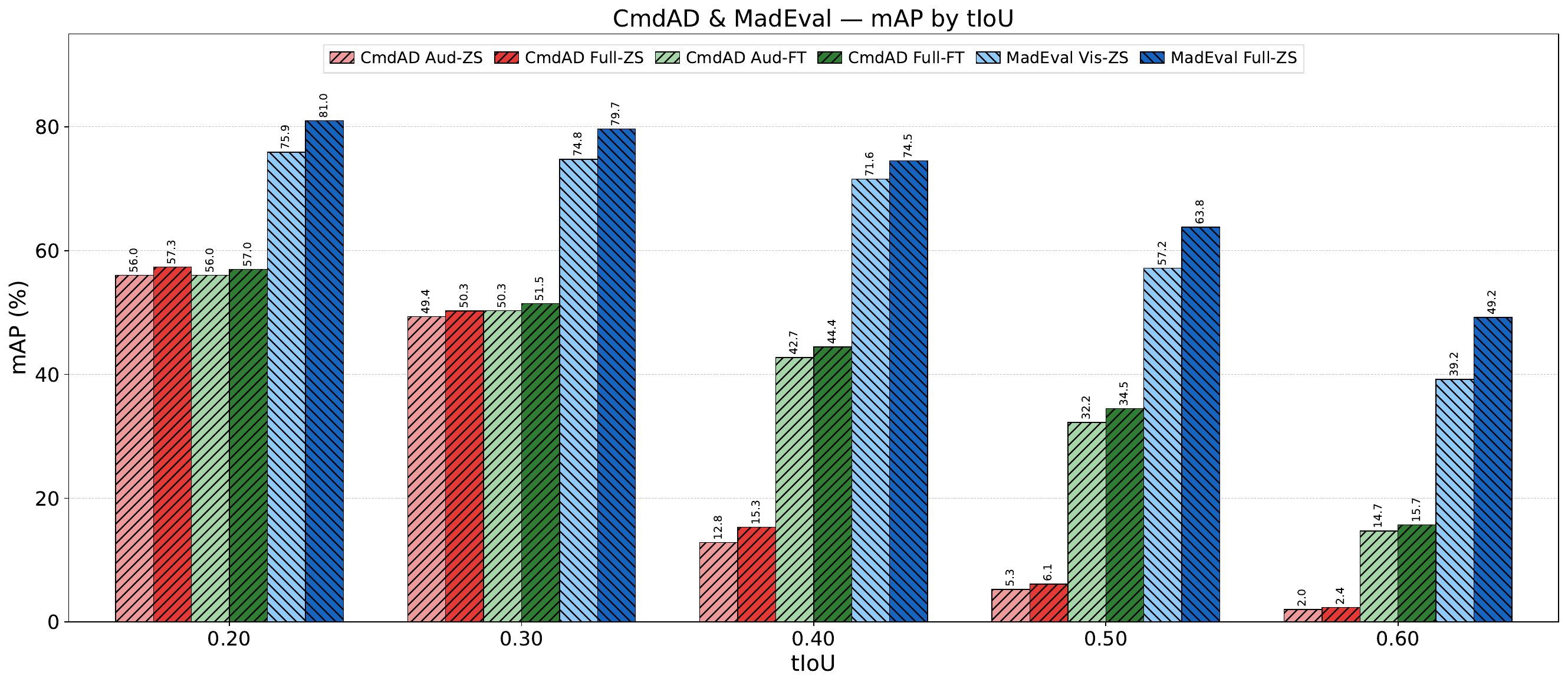}
    \caption{\textbf{Per-tIoU transfer to \cmdad and \madeval.}
    Grouped bar plot of mAP at tIoU thresholds $\{0.2,\,0.3,\,0.4,\,0.5,\,0.6\}$
    on the two transfer benchmarks. \cmdad reports \spoken mAP with
    audio-only (\textbf{Aud}, red) and the full \stageone (\textbf{Full}, dark
    red) under both zero-shot (\textbf{ZS}, hatched) and fine-tuned
    (\textbf{FT}, solid) settings. \madeval reports \visual-window mAP for
    video-only (\textbf{Vis-ZS}, light blue, hatched) and the full \stageone
    (\textbf{Full-ZS}, dark blue, solid), evaluated zero-shot. On \cmdad, ZS
    and FT are nearly identical at loose tIoU but FT pulls sharply ahead at
    stricter thresholds, showing that target-dataset fine-tuning primarily
    sharpens boundary precision. \stageone's lead persists across all
    thresholds and both transfer benchmarks.}
    \label{fig:per_iou_transfer}
\end{figure*}

Figs.~\ref{fig:per_iou_ablation} and~\ref{fig:per_iou_transfer} break down mAP
by tIoU threshold $\{0.2, 0.3, 0.4, 0.5, 0.6\}$. Three patterns emerge.
First, Gemini~2.5~Pro~\cite{comanici2025gemini} is competitive at loose tIoU on \longlsmdc
($\sim 12$~mAP at $0.2$) but collapses sharply as the threshold tightens,
falling below $3$~mAP at tIoU~$0.5$ and effectively zero at $0.6$. This
indicates that prompted timestamp prediction from a general-purpose VLM can
identify approximately when an event occurs but cannot place precise temporal
boundaries — exactly the property AD localization requires.
Second, on \longlsmdc, \stageone's margin over the unimodal baselines is modest
at tIoU~$0.2$ but widens at stricter thresholds: at tIoU~$0.5$, \stageone
reaches $30.4$/$28.6$ vs. video-only's $22.8$/$17.2$ and audio-only's
$21.8$/$25.3$. This is consistent with the modality roles observed elsewhere
in our analysis: audio carries the dominant cue for spoken-slot timing, while
video provides the scene-change and action cues needed to disambiguate which
silence corresponds to a narratable event (Table~\ref{tab:stage1_mAP_by_test}).
Third, the same pattern holds zero-shot on \cmdad and \madeval, where
\stageone keeps its lead across all thresholds. On \cmdad, target-dataset
fine-tuning has little effect at loose tIoU ($\sim 56$ vs. $\sim 57$ at $0.2$)
but produces a large gain at strict tIoU (FT $42.7$ vs.\ ZS $12.8$ at $0.4$),
showing that fine-tuning primarily improves boundary precision rather than
event recall. The dual-head structure transfers asymmetrically by design: the
\spoken head supervises \cmdad-style spoken slots, the \visual head supervises
\madeval-style visual clips, and each head transfers to the benchmark whose
window type it was trained against.

\subsection{Details of the CA3D F1 Comparison on MAD-Eval}
\label{sec:madeval_f1}
The $90.0$ F1 reported for \stageone on \madeval in Sec.~\ref{sec:loc_results} follows the top-1 protocol used by CA3D~\cite{Lee2025CA3D}, which we reproduce here for completeness. \madeval provides exactly one GT \visual per clip. For each clip, we retain only the single highest-scoring predicted segment and compute its temporal IoU against the GT window. At a given tIoU threshold~$\theta$, a clip contributes as follows: no prediction counts as a false negative ($\mathrm{FN}\mathrel{+}=1$); a top-1 prediction with $\mathrm{tIoU} \geq \theta$ counts as a true positive ($\mathrm{TP}\mathrel{+}=1$); a top-1 prediction below the threshold contributes to both false positives and false negatives ($\mathrm{FP}\mathrel{+}=1$, $\mathrm{FN}\mathrel{+}=1$), since the prediction is an incorrect detection and the ground-truth window is simultaneously missed. Precision, recall, and F1 are aggregated across all clips as
\begin{equation}
    P = \frac{\mathrm{TP}}{\mathrm{TP} + \mathrm{FP}}, \quad
    R = \frac{\mathrm{TP}}{\mathrm{TP} + \mathrm{FN}}, \quad
    F_1 = \frac{2 P R}{P + R}.
\end{equation}
Under this accounting, every missed clip contributes symmetrically to $\mathrm{FP}$ and $\mathrm{FN}$, so precision and recall are numerically equal and $F_1$ coincides with both. Note also that the top-1 restriction is conservative for \stageone, since additional correct predictions the model produces on multi-event clips are discarded before scoring; the $90.0$ vs.\ $65.3$ gap therefore reflects the model's boundary precision on its single strongest detection, not any advantage from producing more candidates.

\subsection{UniAD on \longlsmdc}
We evaluate UniAD in the Table~\ref{tab:stage2_generation} setting (GT spoken windows) using the released checkpoint pretrained on MAD, the most closely related available training source to our LSMDC-derived benchmark. Inputs are adapted by converting annotations to clip-relative coordinates and extracting CLIP ViT-L/14 features to match UniAD's expected feature space. UniAD runs without character features, so we compare against our no-CharBank setting. We did not fine-tune UniAD on \longlsmdc; doing so would require train-split conversion to MAD-train format, training-split feature extraction and character-pathway configuration, and we expect it to close part of the reported gap.

\section{Human Evaluation}
\label{sec:human_eval}
\begin{figure}[t]
\centering
\includegraphics[width=\linewidth]{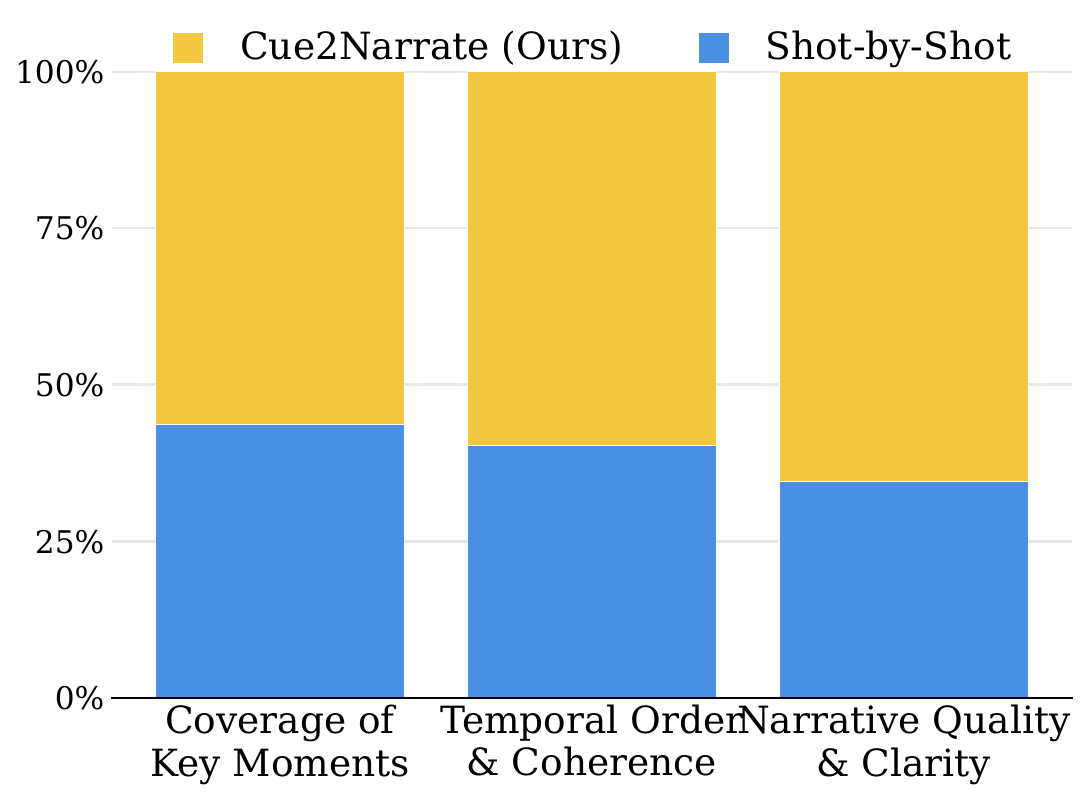}
\caption{\textbf{Human evaluation on \longlsmdc.} Pairwise preference between
\ours (yellow) and Shot-by-Shot~\cite{Xie2025ShotByShot} (blue) across three
dimensions. Evaluators preferred \ours on all three axes, with the largest
margin on Narrative Quality \& Clarity ($\sim$65\% vs.\ 35\%).}
\label{fig:human_eval}
\end{figure}

To complement automatic metrics, we run a pairwise preference study comparing
\ours against Shot-by-Shot~\cite{Xie2025ShotByShot} on $100$ randomly sampled
\longlsmdc clips. For each clip, $20$ evaluators are shown the video alongside
two anonymised AD tracks (ours and the baseline, order randomised) and asked
to choose the preferred track on three axes: (1) \textbf{Coverage of key
moments} whether the ADs cover the salient on-screen actions and
interactions; (2) \textbf{Temporal order \& coherence} whether the ADs
appear at appropriate moments and follow the story flow; and (3)
\textbf{Narrative quality \& clarity} whether the ADs read as coherent,
listener-friendly narration. Fig.~\ref{fig:human_eval} reports the
percentage of pairwise judgments preferring each system. \ours is preferred
on every axis, with the largest gap on narrative quality, consistent with
the AD-style framing the Description Ranking Loss is designed to encourage.
Our evaluators are sighted by design: the three axes require comparing the AD track against the visual ground truth, which is only possible with visual access. This makes the study complementary to, not a substitute for, a listening-experience study with blind and low-vision users; see Sec.~\ref{sec:limitations}.

\section{Examples of negative captions.}
\label{sec:qwen_drl}

To construct the negatives used by DRL (Sec.~\ref{sec:stage2}), we prompt the
frozen Qwen3-VL backbone with a simple prompt
(shown in the box below) on each training clip and cache the resulting
caption offline. The same prompt format is used across all clips, so the
negatives reflect the backbone's default captioning prior long,
descriptive, scene-summarising rather than any AD-style framing. Two
representative pairs of (caption negative, ground-truth AD) are shown in
Fig.~\ref{fig:drl_negatives}; the detailed-vs-concise gap visible in both
examples is what DRL is trained to exploit.

\begin{tcolorbox}[
    colback=gray!5,
    colframe=gray!60,
        fonttitle=\bfseries,
    boxrule=0.6pt,
    arc=2pt,
    left=6pt,
    right=6pt,
    top=6pt,
    bottom=6pt
]
\textbf{System prompt.}
You are an intelligent chatbot designed for generating audio descriptions for movies.
Audio Description (AD) provides concise spoken narration of important visible events,
actions, characters, and scene changes in a movie. 
The prompt includes several short examples of valid AD sentences.

\vspace{0.5em}
\textbf{User prompt.}
\texttt{<video>} Watch the following movie clip and generate an Audio Description for it.

\vspace{0.5em}
\textbf{Constraints.}
The generated AD should be exactly one sentence, use present tense, describe only
visible information, avoid hallucination, focus on important people, actions, and
interactions, and avoid naming characters unless their identity is visually explicit
or provided.
\end{tcolorbox}

\tcbset{
  negex/.style={
    colback=gray!5,
    colframe=gray!40,
    boxrule=0.5pt,
    arc=2pt,
    left=6pt, right=6pt, top=4pt, bottom=4pt,
    fonttitle=\bfseries\small,
    coltitle=black,
    title filled=false,
    fontupper=\footnotesize,
    breakable
  }
}

\begin{figure*}[t]
\centering

\begin{tcolorbox}[negex, title={Example 1 \quad \texttt{train\_3001\_21\_JUMP\_STREET\_clip0\_176.321\_178.664.mp4}}]
\textbf{Caption negative}:
{\itshape The video begins with a close-up shot of a young man with short, light-colored hair, wearing a white shirt. He is sitting on a bench, looking off to the side with a serious expression. The background shows a brick wall and a green bench, suggesting an outdoor setting, possibly a school or public area. The scene then transitions to a wide shot of the exterior of a building, which appears to be a school or institutional facility, with a glass door marked ``School of Business.'' Two individuals are walking away from the entrance.}

\medskip
\textbf{Ground-truth AD:} Sitting on a nearby bench, the nerd looks over.
\end{tcolorbox}

\vspace{2pt}

\begin{tcolorbox}[negex, title={Example 2 \quad \texttt{train\_3001\_21\_JUMP\_STREET\_clip0\_179.061\_182.119.mp4}}]
\textbf{Caption negative}:
{\itshape The video opens with a wide shot of the exterior of a school building, where two students are standing outside. One student, wearing a purple and white jacket, is leaning against the wall with their head down, appearing to be in distress or deep thought. The camera then zooms in on the student, who continues to lean against the wall, occasionally wiping their face with their hand, suggesting they might be crying.}

\medskip
\textbf{Ground-truth AD:} The jock slumps down on a bench on the other side of two columns.
\end{tcolorbox}

\caption{\textbf{Caption negatives used by the Description Ranking Loss.} For each training clip, a frozen LLM Qwen3-VL produces a long, descriptive caption (italicized) that DRL pushes the model away from in favor of the concise ground-truth AD.}
\label{fig:drl_negatives}
\end{figure*}

\section{Gemini prompt details for Action-Score metric}
\label{sec:gemini_prompts}

\paragraph{Reproducibility of the Action Score.} Action Score is an existing metric introduced by Shot-by-Shot; computing it on \longlsmdc requires no access to closed-source software. The closed-source model is used at exactly one point: offline preprocessing of the GT annotations, replacing character names with pronouns and splitting each AD into single-action subsentences (prompts below). We release the resulting GT action annotations alongside the dataset---the same practice as Shot-by-Shot, which ships its precomputed GT action files with its code. Given these files, evaluating any prediction is fully reproducible: the predicted AD is parsed with rule-based dependency parsing (no model calls) and compared to the GT actions with an open-source sentence-embedding model (GTE) plus verb matching. No paid or proprietary model is needed to evaluate on \longlsmdc.

\clearpage

\promptbox{System Prompt}{
\noindent\texttt{<role>}\\[0.5em]
    You are an intelligent chatbot designed for removing character information of a sentence.
\\[0.5em]\noindent\texttt{</role>}
\\\noindent\texttt{<instruction>}\\[0.5em]
    Here ’s how you can accomplish the task :
    You should replace all character information including names , roles , and jobs into pronouns (e.g., he , she , they , her , him , them ).
    Note , objects, locations , and animals are not counted as character information and should be kept as - is .
\\[0.5em]\noindent\texttt{</instruction>}
\\\noindent\texttt{<output\_format>}\\[0.5em]
    You should output the result in JSON format WITHOUT providing ANY more sentences at the beginning or at the end .
\\[0.5em]\noindent\texttt{</output\_format>}
}

\userpromptbox{User Prompt}{
Please read the sentence below that describes a video clip: \\
Input sentence: \texttt{<text\_gt>}\\
Replace all character information including names , roles , and jobs into pronouns (e.g., he , she , they , her , him , them ).
Note , objects, locations , and animals are not counted as character information and should be kept as - is .\\
Here are some examples:
\begin{itemize}
    \item Example1:
    Input sentence : Spicoli watches Mr . Hand pass out the schedule.
    Ouput : He watches him pass out the schedule.
    \item Example 2:
    Input sentence : Waiting for a reply , the inspector has a look of smug satisfaction as he combs his neat moustache.
    Output: Waiting for a reply , he has a look of smug satisfaction as he combs his neat moustache.
    \item Example 3:
    Input sentence : Emmerich ’s eyebrows twitch as he watches her.
    Output : His eyebrows twitch as he watches her.
    \item Example 4:
    Input sentence : Inside is a second pair of doors.
    Output : Inside is a second pair of doors.
    \item Example 5:
    Input sentence : The blonde saunters over to him in her striped pantsuit and leans in for a kiss.
    Output : She saunters over to him in her striped pantsuit and leans in for a kiss.
 \end{itemize}
}

\clearpage

Then the sentence with the removed character is used as input, and the ground truth actions are extracted: 

\promptbox{System Prompt}{
\noindent\texttt{<role>}\\[0.5em]
    You are an intelligent chatbot designed for decompose the sentence into subsentences .
\\[0.5em]\noindent\texttt{</role>}
\\\noindent\texttt{<instruction>}\\[0.5em]
    Here ’s how you can accomplish the task :
    You should split ( rewrite if needed) the sentence into subsentences , each containing only one action phrase (i.e., verb phrase ).
\\[0.5em]\noindent\texttt{</instruction>}
\\\noindent\texttt{<output\_format>}\\[0.5em]
    You should output your answer in JSON format WITHOUT providing ANY more sentences at the beginning or at the end .
\\[0.5em]\noindent\texttt{</output\_format>}
}

\userpromptbox{User Prompt}{
Please read the sentence below that describes a video clip:
Input sentence : \texttt{<text\_wo\_char>}\\
Split and rewrite the sentence into subsentences , each containing only one action (i.e.,verb phrase ) and preserving all other information (e.g., locations , time , affections , etc .) .
Do not output repeating actions.\\
Examples :

\begin{itemize}
    \item Example1:
Input sentence : He watches him pass out the schedule .
Subsentences : [He watches him ., He passes out the schedule .]"
    \item Example2:
Input sentence: Waiting for a reply , he has a look of smug satisfaction as he combs his neat moustache .
Subsentences : [He waits for a reply ., He has a look of smug satisfaction ., He combs his neat moustache .]
    \item Example3:
Input sentence : He swings in front of Kingpin , then bounces off a building and kicks the criminal into the air .
Subsentences : [He swings in front of him ., He bounces off a building ., He kicks him into the air .]
    \item Example4:
Input sentence : His eyebrows twitch as he watches her .
Subsentences : [His eyebrows twitch ., He watches her .]
    \item Example5:
Input sentence : Inside is a second pair of doors .
Subsentences : [Inside is a second pair of doors .]
\end{itemize}
}

\section{\ours (Stage~2) Fine-tuning Prompts}
\label{sec:loc2narrate_prompts}

For fine-tuning the \ours (Stage~2) module, we employ structured prompts to generate audio descriptions conditioned on localized temporal boundaries. Prompts \texttt{System Prompt} and \texttt{User Prompt} show the complete prompt design used during fine-tuning.

The \texttt{System Prompt}  establishes the model's role as an audio description generator and defines the output format requirements, ensuring consistent structure across all generated descriptions. The \texttt{User Prompt} provides the necessary context including video metadata, temporal boundaries from the localization stage, and ground-truth captions to guide the generation process. This two-stage prompting strategy enables the model to learn the mapping from localized temporal segments to natural language audio descriptions.

\promptbox{System Prompt}{
  You are an intelligent chatbot designed for generating audio descriptions for movies.
      Audio description (AD), also referred to as a video description, described video,
  or visual description, is a form of narration used to provide information surrounding
  key visual elements in a media work for the benefit of blind and visually impaired
  consumers.
      Audio descriptions are typically placed within a gap in dialogue or important
  sound.

Here are some examples of audio descriptions that were generated for a different movie clip:
\begin{itemize}[leftmargin=1em, nosep, topsep=2pt]
  \item Example 1: Sitting on a nearby bench, the nerd looks over.      
  \item Example 2: Now students spar in a hand-to-hand combat class.
  \item Example 3: He slams him down on the mat and raises his fist.
  \item Example 4: She stares distantly, then faces Andy.
  \item Example 5: A young girl cycles through the city at night.
\end{itemize}
}

\userpromptbox{User Prompt}{
\noindent\texttt{<video>}\\[0.5em]
Watch the following movie clip and generate an Audio Description for it.
      The Audio Description should only consist of one sentence.
      Make sure you do not hallucinate information and only describe what you can see in
  the movie clip.

      Requirements:

\begin{itemize}
    \item Describe only the most important visible information in the clip.

    \item Focus on the visible people or subjects, their actions, and important
  interactions.

    \item Use present tense.

    \item Do not name characters unless their identity is explicitly provided in the input
  or visually confirmed. Instead use neutral words such as "A man", "A woman", "He",
  "She" etc.   Assistant target
the GT AD sentence, e.g.: Sitting on a nearby bench, the nerd looks over.
\end{itemize}
}

\label{sec:stage1_qual_analysis}

\section{Additional Qualitative Results}
\label{sec:qual_both}

\subsection{Systematic error categories}
\begin{figure*}[t]
     \begin{tikzpicture}
        \node[anchor=south west, inner sep=0] (img1) at (0,0)
            {\includegraphics[width=\linewidth]{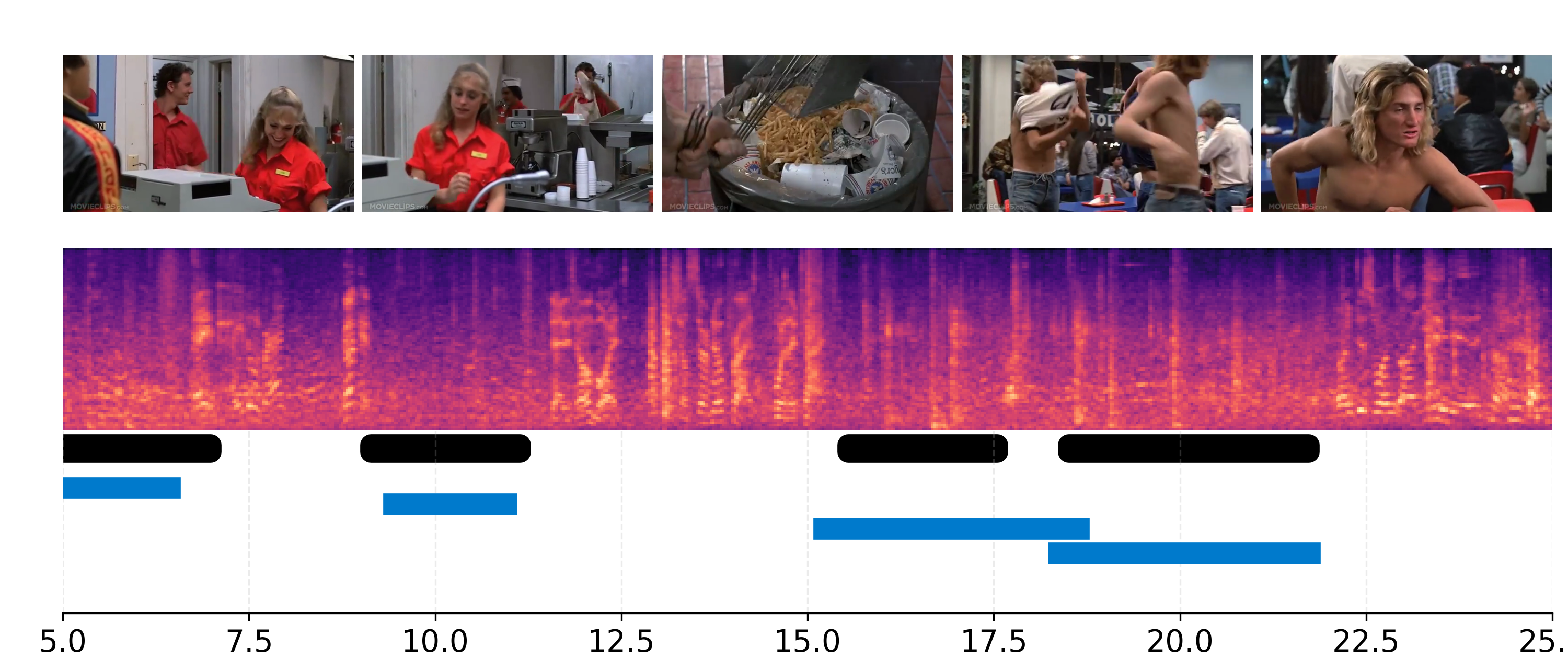}};
                \begin{scope}[x={(img1.south east)}, y={(img1.north west)}]
            \node[anchor=north west,
                  font=\scriptsize,
                  text=white,
                                    fill opacity=0.6,
                  text opacity=1,
                  inner sep=2pt]
                at (0.07,0.34) {GT 1};
            \node[anchor=north west,
                  font=\scriptsize,
                  text=white,
                                    fill opacity=0.6,
                  text opacity=1,
                  inner sep=2pt]
                at (0.27,0.34) {GT 2};
            \node[anchor=north west,
                  font=\scriptsize,
                  text=white,
                                    fill opacity=0.6,
                  text opacity=1,
                  inner sep=2pt]
                at (0.57,0.34) {GT 3};
            \node[anchor=north west,
                  font=\scriptsize,
                  text=white,
                                    fill opacity=0.6,
                  text opacity=1,
                  inner sep=2pt]
                at (0.73,0.34) {GT 4};
            \node[anchor=north west,
                  font=\scriptsize,
                  text=white,
                                    fill opacity=0.6,
                  text opacity=1,
                  inner sep=2pt]
                at (0.06,0.28) {Pred 1};
            \node[anchor=north west,
                  font=\scriptsize,
                  text=white,
                                    fill opacity=0.6,
                  text opacity=1,
                  inner sep=2pt]
                at (0.264,0.258) {Pred 2};
            \node[anchor=north west,
                  font=\scriptsize,
                  text=white,
                                    fill opacity=0.6,
                  text opacity=1,
                  inner sep=2pt]
                at (0.57,0.22) {Pred 3};
            \node[anchor=north west,
                  font=\scriptsize,
                  text=white,
                                    fill opacity=0.6,
                  text opacity=1,
                  inner sep=2pt]
                at (0.73,0.18) {Pred 4};
        
        \end{scope}
    \end{tikzpicture}

                    \caption{\textbf{CMD-AD qualitative examples on curated short clips.} Our model demonstrates effective localization and generation on moderately dense sequences. (Top) Sampled video frames, (Middle) mel-spectrogram showing audio activity (bright regions) and silence periods (dark regions), (Bottom) temporal alignment with ground-truth segments (black bars) and predicted segments (blue bars). The model successfully identifies appropriate silence windows and generates contextually relevant descriptions. GT: Ground Truth, Pred: Prediction.}
    \label{fig:cmdad_qual1}
\end{figure*}
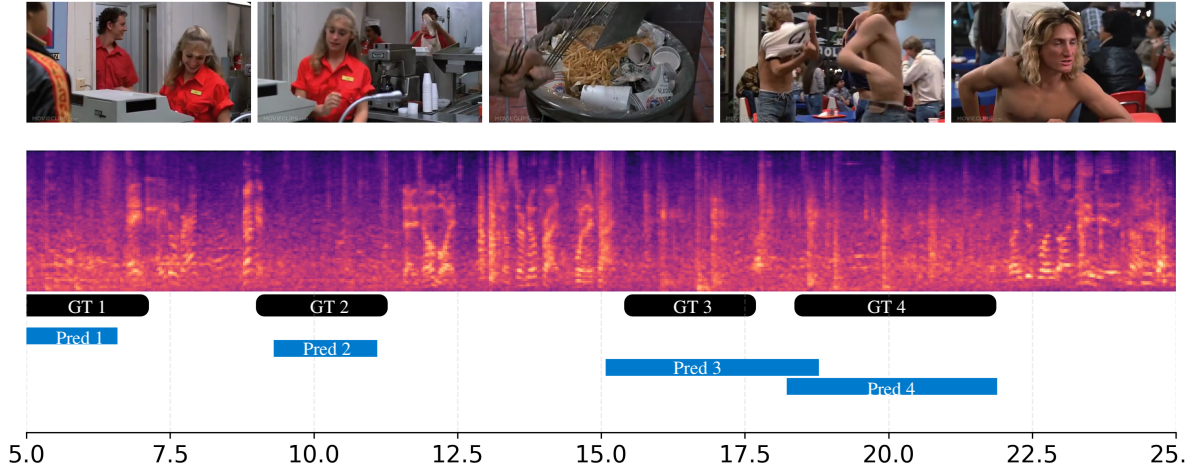

\begin{figure*}[t]
    \centering
        \begin{tikzpicture}
        \node[anchor=south west, inner sep=0] (img3) at (0,0)
            {\includegraphics[width=\linewidth]{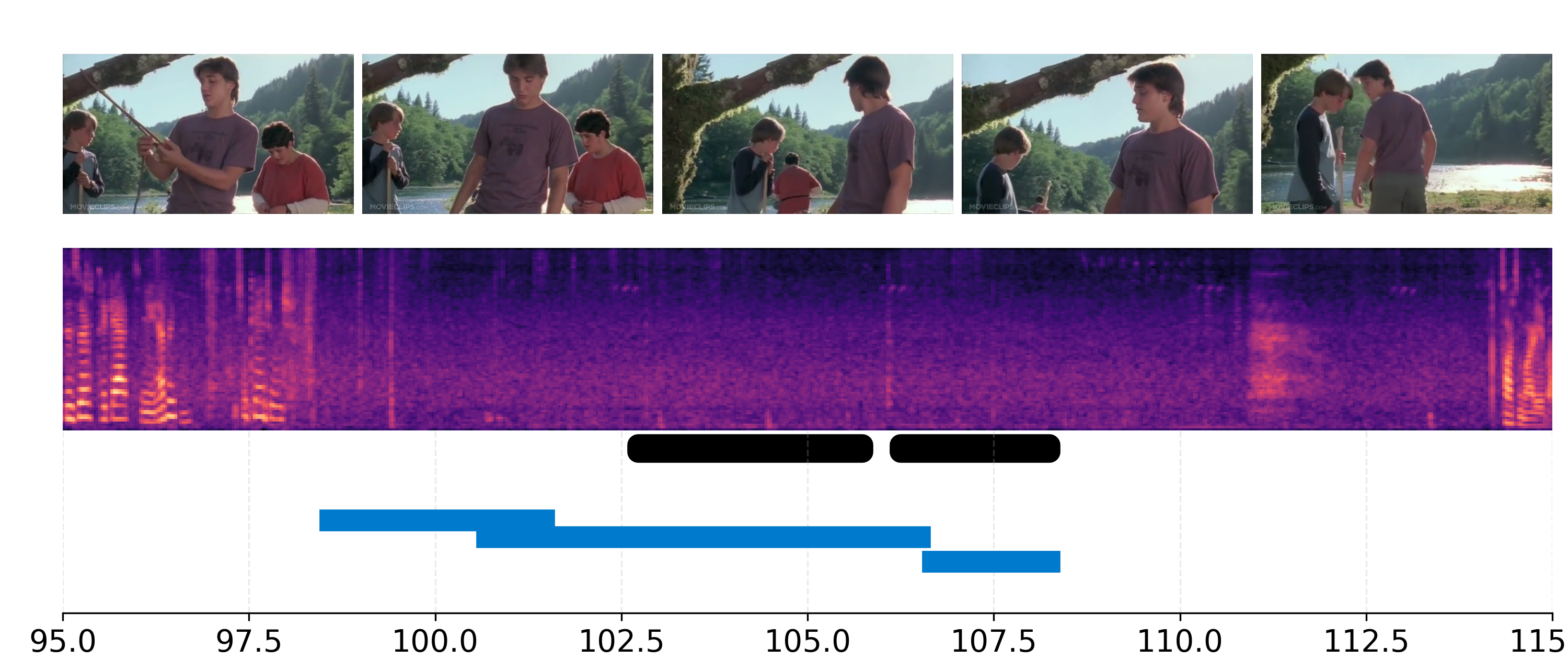}};
        \begin{scope}[x={(img3.south east)}, y={(img3.north west)}]
            \node[anchor=north west,
                  font=\scriptsize,
                  text=white,
                                    fill opacity=0.6,
                  text opacity=1,
                  inner sep=2pt]
                at (0.46,0.34) {GT 1};
            \node[anchor=north west,
                  font=\scriptsize,
                  text=white,
                                    fill opacity=0.6,
                  text opacity=1,
                  inner sep=2pt]
                at (0.6,0.34) {GT 2};
            \node[anchor=north west,
                  font=\scriptsize,
                  text=white,
                                    fill opacity=0.6,
                  text opacity=1,
                  inner sep=2pt]
                at (0.25,0.23) {Pred 1};
            \node[anchor=north west,
                  font=\scriptsize,
                  text=white,
                                    fill opacity=0.6,
                  text opacity=1,
                  inner sep=2pt]
                at (0.4,0.205) {Pred 2};
            \node[anchor=north west,
                  font=\scriptsize,
                  text=white,
                                    fill opacity=0.6,
                  text opacity=1,
                  inner sep=2pt]
                at (0.62,0.17) {Pred 3};
        \end{scope}
    \end{tikzpicture}
    \noindent\makebox[\linewidth][l]{\footnotesize GT 1: George packs up his Swiss Army knife and plods away.}
            \noindent\makebox[\linewidth][l]{\footnotesize Pred 2: George folds up his knife and slowly walks away.}

    \noindent\makebox[\linewidth][l]{\footnotesize GT 2: Sam and Rocky watch him walking away.}
        \noindent\makebox[\linewidth][l]{\footnotesize Pred 3: He stands on one leg, then the other.}
    
          \begin{tikzpicture}
        \node[anchor=south west, inner sep=0] (img3) at (0,0)
            {\includegraphics[width=\linewidth]{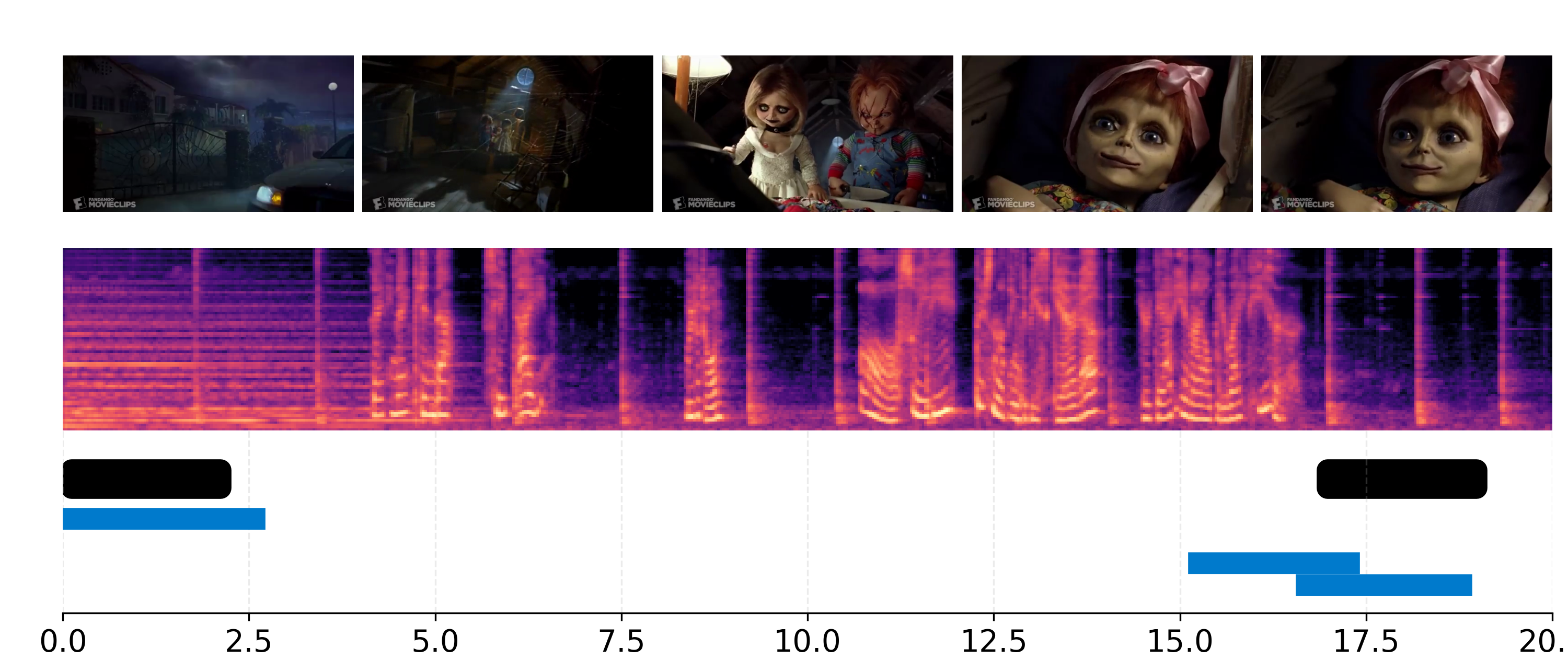}};
        \begin{scope}[x={(img3.south east)}, y={(img3.north west)}]
            \node[anchor=north west,
                  font=\scriptsize,
                  text=white,
                                    fill opacity=0.6,
                  text opacity=1,
                  inner sep=2pt]
                at (0.07,0.3) {GT 1};
            \node[anchor=north west,
                  font=\scriptsize,
                  text=white,
                                    fill opacity=0.6,
                  text opacity=1,
                  inner sep=2pt]
                at (0.88,0.3) {GT 2};
            \node[anchor=north west,
                  font=\scriptsize,
                  text=white,
                                    fill opacity=0.6,
                  text opacity=1,
                  inner sep=2pt]
                at (0.07,0.23) {Pred 1};
            \node[anchor=north west,
                  font=\scriptsize,
                  text=white,
                                    fill opacity=0.6,
                  text opacity=1,
                  inner sep=2pt]
                at (0.8,0.17) {Pred 2};
            \node[anchor=north west,
                  font=\scriptsize,
                  text=white,
                                    fill opacity=0.6,
                  text opacity=1,
                  inner sep=2pt]
                at (0.86,0.13) {Pred 3};
        \end{scope}
    \end{tikzpicture}
    \noindent\makebox[\linewidth][l]{\footnotesize GT 1: Tiffany reaches for the lamp.}
    \noindent\makebox[\linewidth][l]{\footnotesize Pred 1: The light shines through the window above the attic door. }
        \noindent\makebox[\linewidth][l]{\footnotesize GT 2: Glen, Glenda stares at Chucky.}
    \noindent\makebox[\linewidth][l]{\footnotesize Pred 2: Tiffany and Chucky look at each other.}
        \caption{\textbf{CMD-AD example demonstrating selective silence awareness.} The model learns that not every acoustic silence requires an audio description. In the top example, multiple silence gaps appear in the mel-spectrogram throughout the 25-second clip, yet the model correctly predicts only segments corresponding to visually salient events that align with ground truth. This demonstrates that audio-visual fusion enables discrimination between acoustically silent moments that are narratively important versus those that are not. The visual stream determines content salience while audio detects insertion opportunities. GT: Ground Truth, Pred: Prediction.}
    \label{fig:cmdad_qual2}
\end{figure*}

\begin{figure*}[t]
     \begin{tikzpicture}
        \node[anchor=south west, inner sep=0] (img1) at (0,0)
            {\includegraphics[width=\linewidth]{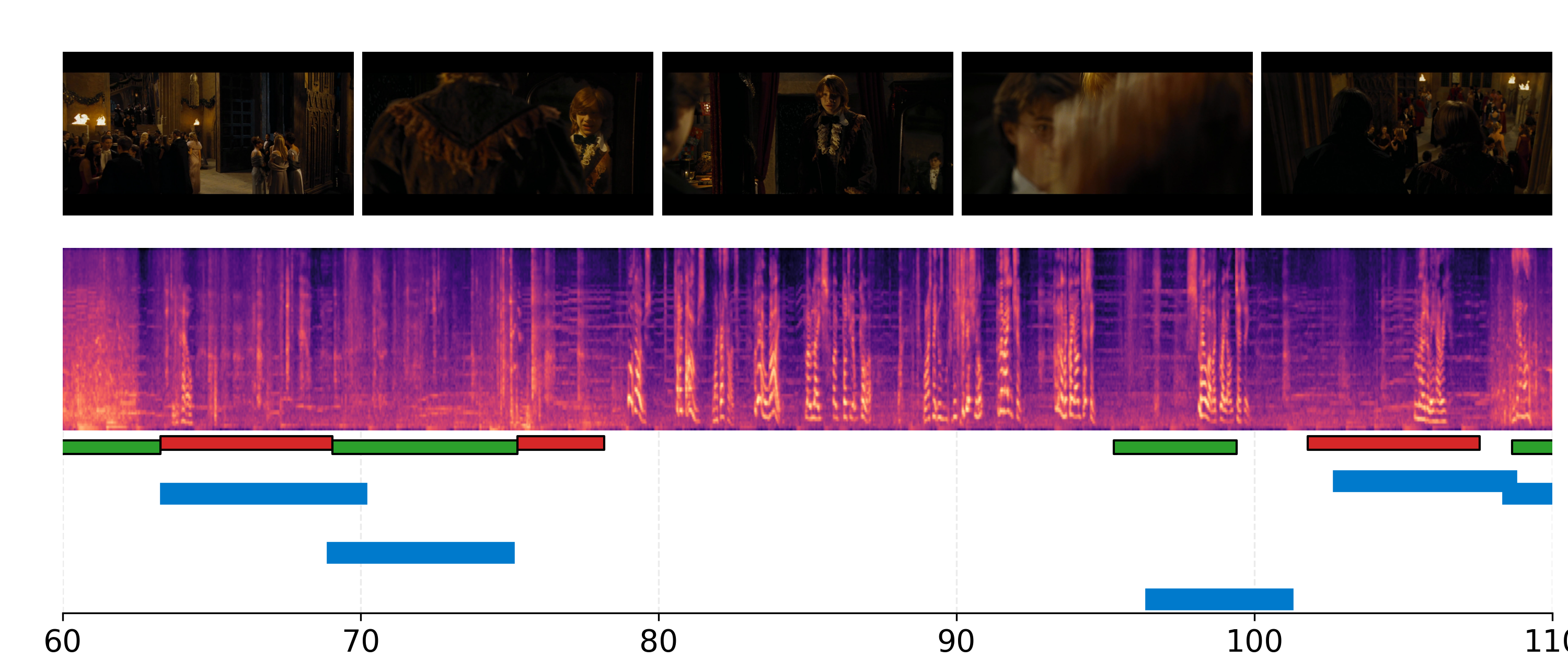}};
                \begin{scope}[x={(img1.south east)}, y={(img1.north west)}]
            \node[anchor=north west,
                  font=\scriptsize,
                  text=white,
                  fill=black,
                  fill opacity=0.6,
                  text opacity=1,
                  inner sep=2pt]
                at (0.05,0.34) {GT 1};
            \node[anchor=north west,
                  font=\scriptsize,
                  text=white,
                  fill=black,
                  fill opacity=0.6,
                  text opacity=1,
                  inner sep=2pt]
                at (0.14,0.34) {GT 2};
            \node[anchor=north west,
                  font=\scriptsize,
                  text=white,
                  fill=black,
                  fill opacity=0.6,
                  text opacity=1,
                  inner sep=2pt]
                at (0.24,0.34) {GT 3};
            \node[anchor=north west,
                  font=\scriptsize,
                  text=white,
                  fill=black,
                  fill opacity=0.6,
                  text opacity=1,
                  inner sep=2pt]
                at (0.34,0.34) {GT 4};
            \node[anchor=north west,
                  font=\scriptsize,
                  text=white,
                  fill=black,
                  fill opacity=0.6,
                  text opacity=1,
                  inner sep=2pt]
                at (0.73,0.34) {GT 5};
            \node[anchor=north west,
                  font=\scriptsize,
                  text=white,
                  fill=black,
                  fill opacity=0.6,
                  text opacity=1,
                  inner sep=2pt]
                at (0.86,0.34) {GT 6};
            \node[anchor=north west,
                  font=\scriptsize,
                  text=white,
                  fill=black,
                  fill opacity=0.6,
                  text opacity=1,
                  inner sep=2pt]
                at (0.94,0.34) {GT 7};
            \node[anchor=north west,
                  font=\scriptsize,
                  text=white,
                  fill=black,
                  fill opacity=0.6,
                  text opacity=1,
                  inner sep=2pt]
                at (0.16,0.27) {Pred 1};
            \node[anchor=north west,
                  font=\scriptsize,
                  text=white,
                  fill=black,
                  fill opacity=0.6,
                  text opacity=1,
                  inner sep=2pt]
                at (0.26,0.17) {Pred 2};
            \node[anchor=north west,
                  font=\scriptsize,
                  text=white,
                  fill=black,
                  fill opacity=0.6,
                  text opacity=1,
                  inner sep=2pt]
                at (0.75,0.12) {Pred 3};
            \node[anchor=north west,
                  font=\scriptsize,
                  text=white,
                  fill=black,
                  fill opacity=0.6,
                  text opacity=1,
                  inner sep=2pt]
                at (0.88,0.29) {Pred 4};
            \node[anchor=north west,
                  font=\scriptsize,
                  text=white,
                  fill=black,
                  fill opacity=0.6,
                  text opacity=1,
                  inner sep=2pt]
                at (0.96,0.26) {Pred 5};        
        \end{scope}
    \end{tikzpicture}\\[1mm]
    \noindent\makebox[\linewidth][l]{\footnotesize GT 1: The girls wear long gowns, the boys dress robes.}
    \noindent\makebox[\linewidth][l]{\footnotesize GT 2: In his room, Ron adjusts his tatty robe.}
    \noindent\makebox[\linewidth][l]{\footnotesize Pred 1: HARRY is in the Great Hall, at the Gryffindor table. }
    \noindent\makebox[\linewidth][l]{\footnotesize GT 3: Ron stares at himself in a mirror. }
    \noindent\makebox[\linewidth][l]{\footnotesize Pred 2: The queen turns to her son. }
    \noindent\makebox[\linewidth][l]{\footnotesize GT 4: Harry comes in wearing a smart black robe and a white bow tie.}
    \noindent\makebox[\linewidth][l]{\footnotesize GT 5: Harry takes a deep breath and sinks down after it.}
    \noindent\makebox[\linewidth][l]{\footnotesize Pred 3: They look at each other. }
    \noindent\makebox[\linewidth][l]{\footnotesize GT 6: Moody searches the shelves.}
    \noindent\makebox[\linewidth][l]{\footnotesize Pred 4: They are all wearing costumes. }
    \noindent\makebox[\linewidth][l]{\footnotesize GT 7: Snowflakes whirl outside the boys' bedroom.}
        \noindent\makebox[\linewidth][l]{\footnotesize Pred 5: They both wear bow ties as it snows.}
                                                                                                                                                                                                                                                                                                                                                                                                                                                                                                                                                          \caption{\textbf{\longlsmdc example showing dense annotations in real-world scenarios.} LongLSMDC presents significantly more challenging scenarios with high annotation density and longer, untrimmed clips. This example shows 7 ground-truth segments within a 50-second window, demonstrating rapid succession of events typical of continuous movie sequences. The model maintains temporal precision despite tight spacing, with predicted segments closely tracking ground truth. The mel-spectrogram reveals sustained silence periods that must be partitioned into multiple discrete AD opportunities based on visual content changes, demonstrating the model's ability to handle continuous narrative coverage. GT: Ground Truth, Pred: Prediction.}
     \label{fig:longlsmdc_qual1}

\end{figure*}

\begin{figure*}[t]
    \centering
                                                                                                                                                                                                                                                                                                                                                                                                                                                                                                                                                                                          \begin{tikzpicture}
        \node[anchor=south west, inner sep=0] (img3) at (0,0)
            {\includegraphics[width=\linewidth]{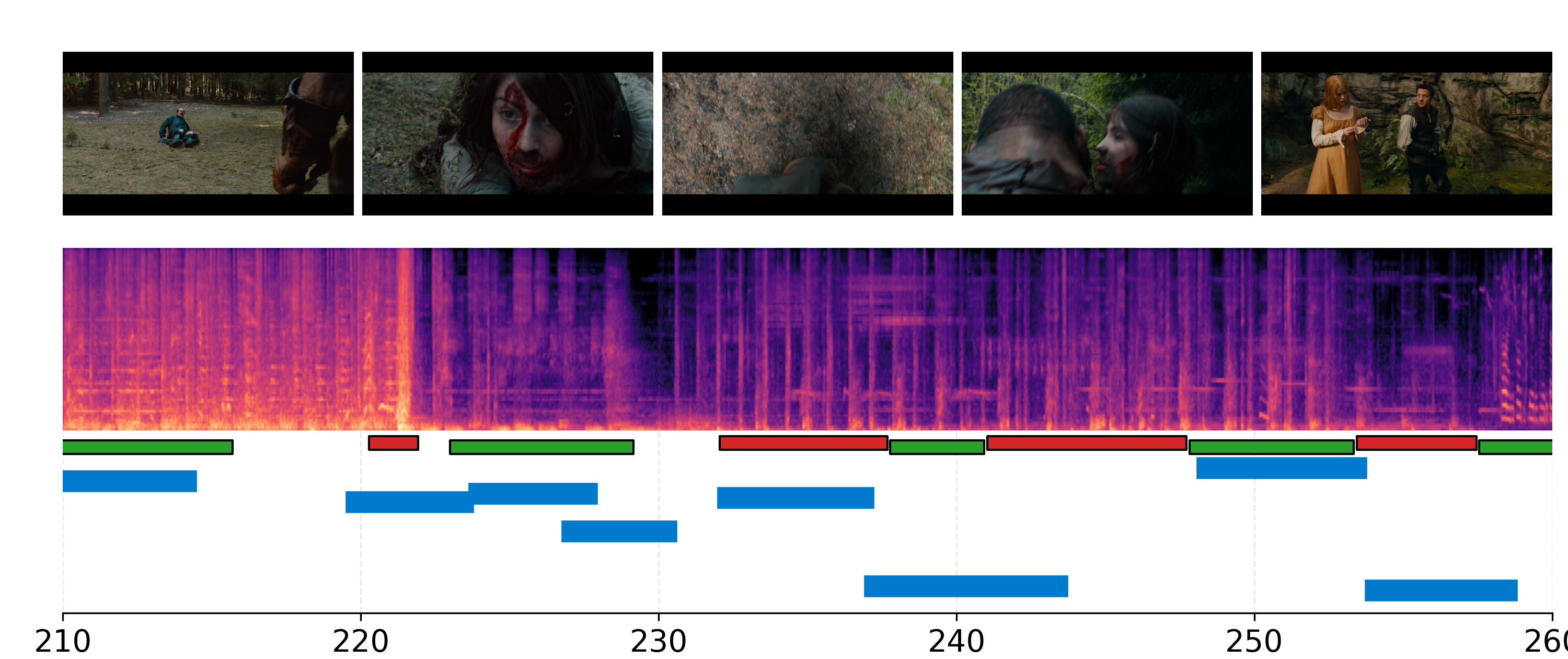}};
        \begin{scope}[x={(img3.south east)}, y={(img3.north west)}]
            \node[anchor=north west,
                  font=\scriptsize,
                  text=white,
                  fill=black,
                  fill opacity=0.6,
                  text opacity=1,
                  inner sep=2pt]
                at (0.08,0.34) {GT 1};
            \node[anchor=north west,
                  font=\scriptsize,
                  text=white,
                  fill=black,
                  fill opacity=0.6,
                  text opacity=1,
                  inner sep=2pt]
                at (0.23,0.34) {GT 2};
            \node[anchor=north west,
                  font=\scriptsize,
                  text=white,
                  fill=black,
                  fill opacity=0.6,
                  text opacity=1,
                  inner sep=2pt]
                at (0.34,0.34) {GT 3};
            \node[anchor=north west,
                  font=\scriptsize,
                  text=white,
                  fill=black,
                  fill opacity=0.6,
                  text opacity=1,
                  inner sep=2pt]
                at (0.48,0.34) {GT 4};
            \node[anchor=north west,
                  font=\scriptsize,
                  text=white,
                  fill=black,
                  fill opacity=0.6,
                  text opacity=1,
                  inner sep=2pt]
                at (0.58,0.34) {GT 5};
            \node[anchor=north west,
                  font=\scriptsize,
                  text=white,
                  fill=black,
                  fill opacity=0.6,
                  text opacity=1,
                  inner sep=2pt]
                at (0.66,0.34) {GT 6};
            \node[anchor=north west,
                  font=\scriptsize,
                  text=white,
                  fill=black,
                  fill opacity=0.6,
                  text opacity=1,
                  inner sep=2pt]
                at (0.8,0.34) {GT 7};
            \node[anchor=north west,
                  font=\scriptsize,
                  text=white,
                  fill=black,
                  fill opacity=0.6,
                  text opacity=1,
                  inner sep=2pt]
                at (0.88,0.34) {GT 8};
            \node[anchor=north west,
                  font=\scriptsize,
                  text=white,
                  fill=black,
                  fill opacity=0.6,
                  text opacity=1,
                  inner sep=2pt]
                at (0.95,0.34) {GT 9};
            \node[anchor=north west,
                  font=\scriptsize,
                  text=white,
                  fill=black,
                  fill opacity=0.6,
                  text opacity=1,
                  inner sep=2pt]
                at (0.06,0.28) {Pred 1};
            \node[anchor=north west,
                  font=\scriptsize,
                  text=white,
                  fill=black,
                  fill opacity=0.6,
                  text opacity=1,
                  inner sep=2pt]
                at (0.24,0.26) {Pred 2};
            \node[anchor=north west,
                  font=\scriptsize,
                  text=white,
                  fill=black,
                  fill opacity=0.6,
                  text opacity=1,
                  inner sep=2pt]
                at (0.32,0.28) {Pred 3};
            \node[anchor=north west,
                  font=\scriptsize,
                  text=white,
                  fill=black,
                  fill opacity=0.6,
                  text opacity=1,
                  inner sep=2pt]
                at (0.37,0.21) {Pred 4};
            \node[anchor=north west,
                  font=\scriptsize,
                  text=white,
                  fill=black,
                  fill opacity=0.6,
                  text opacity=1,
                  inner sep=2pt]
                at (0.49,0.26) {Pred 5};
            \node[anchor=north west,
                  font=\scriptsize,
                  text=white,
                  fill=black,
                  fill opacity=0.6,
                  text opacity=1,
                  inner sep=2pt]
                at (0.59,0.14) {Pred 6};
            \node[anchor=north west,
                  font=\scriptsize,
                  text=white,
                  fill=black,
                  fill opacity=0.6,
                  text opacity=1,
                  inner sep=2pt]
                at (0.8,0.31) {Pred 7};
            \node[anchor=north west,
                  font=\scriptsize,
                  text=white,
                  fill=black,
                  fill opacity=0.6,
                  text opacity=1,
                  inner sep=2pt]
                at (0.9,0.13) {Pred 8};
        \end{scope}
    \end{tikzpicture}
    \noindent\makebox[\linewidth][l]{\footnotesize GT 1: Mina stands with her wand ready after assaulting Muriel with a spell.}
    \noindent\makebox[\linewidth][l]{\footnotesize Pred 1: A man with long hair and a beard, dressed in leather, and carrying a spear, lifts a woman into his arms.}
    \noindent\makebox[\linewidth][l]{\footnotesize GT 2: The troll stomps on his head, his brains and blood splattering on the ground.}
    \noindent\makebox[\linewidth][l]{\footnotesize Pred 2: }
    \noindent\makebox[\linewidth][l]{\footnotesize GT 3: Mina shatters a tree behind Muriel. }
    \noindent\makebox[\linewidth][l]{\footnotesize Pred 3: The man stands in the middle of a clearing.}
    \noindent\makebox[\linewidth][l]{\footnotesize Pred 4: The giant turns and sees Jane. }
    \noindent\makebox[\linewidth][l]{\footnotesize GT 4: Gretel awakens to the gentle rocking motion of being carried. }
    \noindent\makebox[\linewidth][l]{\footnotesize Pred 5: The light from the flashlight beam moves about the ground. }
    \noindent\makebox[\linewidth][l]{\footnotesize GT 6: Gretel turns her head to find the side of the troll's head.}
    \noindent\makebox[\linewidth][l]{\footnotesize GT 5: Ben watches Gretel fall. }
    \noindent\makebox[\linewidth][l]{\footnotesize Pred 6: John reaches out to grab the giant's foot. }
        \noindent\makebox[\linewidth][l]{\footnotesize GT 7: Gretel closes her eyes and leans her head to rest against the giant's shoulder.  }
    \noindent\makebox[\linewidth][l]{\footnotesize Pred 7: A man with long hair and a beard, dressed in leather, and carrying a spear, lifts a woman into his arms. }
        \noindent\makebox[\linewidth][l]{\footnotesize GT 8: They kiss. }
    \noindent\makebox[\linewidth][l]{\footnotesize Pred 8: Muriel and her other witch, one with red hair, hover menacingly. }
                \caption{\textbf{\longlsmdc example showing dense annotation and real-world annotation scenarios.} This sequence contains 9 ground-truth segments within a 50-second window (4-5 adjacent segments in rapid succession around timestamps 210-260s), representing the most challenging high-density scenarios in real-world untrimmed movies. The model must handle: (1) minimal gaps between consecutive ADs, (2) complex visual scenes with multiple characters and actions, (3) extended dialogue-free sequences requiring continuous narrative coverage. Despite these challenges, the localizer successfully partitions extended silence periods into appropriate description units. Some prediction errors occur due to the extreme difficulty of this scenario, highlighting areas for future improvement. GT: Ground Truth, Pred: Prediction.}

    \label{fig:longlsmdc_qual2}
\end{figure*}

We organise generation errors into five categories: wrong-subject attribution, an under-specified verb despite identifying the correct event, boundary drift that describes adjacent content, hallucinated visual detail, and correct content expressed with wording that differs from the reference. The main-paper example ``Ace stands and watches Chris walk away'' versus ``Gordy pulls the hammer'' illustrates wrong-subject attribution when several plausible visual subjects share the frame. Figures~\ref{fig:cmdad_qual1}--\ref{fig:longlsmdc_qual2} provide representative boundary-drift, under-specification, hallucination, and alternative-phrasing cases. This variability is consistent with the inter-annotator variability of AD reported in prior work.

Figures~\ref{fig:cmdad_qual1} and~\ref{fig:cmdad_qual2} illustrate our model's localization and generation performance on \cmdad test clips, while Figures~\ref{fig:longlsmdc_qual1} and~\ref{fig:longlsmdc_qual2} show results on \longlsmdc. Each example shows: (top) sampled video frames from the continuous clip, (middle) mel-spectrogram visualization where bright regions indicate audio activity and dark regions indicate silence periods suitable for AD insertion, and (Bottom) temporal alignment where black bars represent ground-truth AD segments and blue bars show our predicted segments.

\textbf{\cmdad Results: Learning Selective Silence Awareness.} Figure~\ref{fig:cmdad_qual2} demonstrates that our proposed localization approach learns not every silence period requires an AD. In the top example, the mel-spectrogram shows multiple silence gaps (dark regions) throughout the 25-second clip, yet our model correctly predicts only 4 AD segments that align with ground truth. This selectivity reveals that the model distinguishes between acoustically silent moments that are narratively important versus those that are not. The audio-visual fusion enables this discrimination while audio detects silence periods, the visual stream determines whether the corresponding visual content is salient enough to describe. For instance, brief pauses during continuous action may be acoustically silent but visually redundant, requiring no narration. The predicted segments (blue bars) consistently align with both mel-spectrogram silence regions (dark bands) and ground-truth annotations (black bars), confirming that our approach successfully learns the joint constraint of acoustic availability and visual salience.

\textbf{\longlsmdc Results: Dense, Continuous Annotations.} Figures~\ref{fig:longlsmdc_qual1} and~\ref{fig:longlsmdc_qual2} showcase performance on densely annotated sequences where ADs occur in rapid succession, reflecting real-world untrimmed movie scenarios. Figure~\ref{fig:longlsmdc_qual2} is particularly challenging: 4-5 adjacent ground-truth AD segments occur within a 20-second window (timestamps 210-260s), with minimal gaps between them. This represents the high-density scenario characteristic of \longlsmdc, where ADs must be inserted continuously during extended dialogue-free sequences. Our model maintains temporal precision despite this tight spacing, with predicted segments (blue bars) closely tracking ground truth across the rapid sequence. The mel-spectrogram reveals sustained low-activity regions (dark purple bands) spanning multiple seconds, and our proposed localization approach successfully partitions these extended silence periods into multiple discrete AD segments based on visual changes. This demonstrates robust generalization to the challenging case of continuous narrative coverage where the model must identify not just isolated silence gaps, but also how to temporally segment extended silent periods into appropriate segments. Some prediction errors occur due to the extreme difficulty of these scenarios, highlighting the challenges of dense, real-world movie description tasks.

\section{Artifact License}
\label{sec:artifact}
\begin{table}[t]
\begingroup
\centering
\small
\caption{\textbf{Artifact release plan.}}
\resizebox{\columnwidth}{!}{\begin{tabular}{lll}
\toprule
Artifact & Access & Terms\\
\midrule
\longlsmdc annotations (paired windows, splits) & Request & LSMDC research-only\\
Preprocessed clips + extracted features & Request & LSMDC research-only\\
Per-clip character banks & Request & LSMDC research-only\\
GT action annotations (Action Score) & Request & LSMDC research-only\\
Stage~1 + Stage~2 training/inference code & Public & Open license\\
Trained checkpoints (localizer, LoRA adapters) & Public & Open license\\
Evaluation scripts (matching, all metrics) & Public & Open license\\
\bottomrule
\end{tabular}}
\endgroup
\end{table}
The datasets and models employed in this work are subject to the following licenses.
The Qwen2.5 and Qwen3 model families are released under the Apache 2.0 License,
which permits free use, modification, and distribution for both research and commercial
purposes. The Condensed Movies Dataset (CMD) is distributed
under the Creative Commons Attribution 4.0 International License (CC~BY~4.0),
which allows unrestricted use provided appropriate credit is given.
The Large Scale Movie Description Challenge (LSMDC) dataset is made available
under a restricted-access agreement through the Max Planck Institute for Informatics
(MPII), permitting use exclusively for non-commercial academic research.
The MAD-Eval (Movie Audio Descriptions) dataset is governed by a Non-Disclosure Agreement
(NDA) that limits its use to academic research purposes only.
All resources are therefore used in full compliance with their respective licensing
terms within the scope of this academic work.

\section{Model size and budget}
\label{sec:model_size}

\paragraph{Stage 1.} The audio-visual dual-head localizer has 6.64M parameters in total, shared across all three variants (\textit{video-only}, \textit{audio-only}, \textit{full}); mode flags disable compute paths but do not change parameter count.

\paragraph{Stage 2.} We use Qwen2.5-VL-7B or Qwen3-VL-8B as the backbone, adapted with LoRA ($r{=}16$, $\alpha{=}32$, dropout 0.05) on the language model and multimodal projection; the vision encoder is frozen.

\end{document}